\documentclass[journal,twoside,web]{ieeetran}
\usepackage{graphicx}
\usepackage[caption=false,font=footnotesize]{subfig}
\usepackage{subfiles} % Best loaded last in the preamble
\usepackage{multicol}
\usepackage{cite}
\usepackage{amsmath,amssymb,amsfonts}
\usepackage{algorithmic}
\usepackage{graphicx}
\usepackage{textcomp}
\usepackage{hyperref}
\usepackage{xr}
\usepackage[caption=false,font=footnotesize]{subfig}

\begin{document}

\def\BibTeX{{\rm B\kern-.05em{\sc i\kern-.025em b}\kern-.08em
    T\kern-.1667em\lower.7ex\hbox{E}\kern-.125emX}}
% \markboth{\journalname, VOL. XX, NO. XX, XXXX 2021}
% {Lee \MakeLowercase{\textit{et al.}}: Preparation of Papers for IEEE TRANSACTIONS and JOURNALS (July 2021)}

% Colored comment
% \usepackage{xcolor}
% \newcommand{\kisuk}[1]{{\color{orange}{{#1}}}}

%%% HELPER CODE FOR DEALING WITH EXTERNAL REFERENCES
\makeatletter
\newcommand*{\addFileDependency}[1]{
  \typeout{(#1)}
  \@addtofilelist{#1}
  \IfFileExists{#1}{}{\typeout{No file #1.}}
}
\makeatother

\newcommand*{\myexternaldocument}[1]{
    \externaldocument{#1}
    \addFileDependency{#1.tex}
    \addFileDependency{#1.aux}
}
%%% END HELPER CODE

% put all the external documents here!
% \myexternaldocument{supp_fig}

% just to see what's happening
% \listfiles

% \begin{document}

\title{Learning and Segmenting Dense Voxel Embeddings for 3D Neuron Reconstruction}

\author{Kisuk Lee, Ran Lu, Kyle Luther, and H. Sebastian Seung

\thanks{This research was supported by the Intelligence Advanced Research Projects Activity (IARPA) via Department of Interior/ Interior Business Center (DoI/IBC) contract number D16PC0005, NIH/NIMH (U01MH114824, U01MH117072, RF1MH117815), NIH/NINDS (U19NS104648, R01NS104926), NIH/NEI (R01EY027036), and ARO (W911NF-12-1-0594), and the Mathers Foundation. 
%The U.S. Government is authorized to reproduce and distribute reprints for Governmental purposes notwithstanding any copyright annotation thereon. Disclaimer: The views and conclusions contained herein are those of the authors and should not be interpreted as necessarily representing the official policies or endorsements, either expressed or implied, of IARPA, DoI/IBC, or the U.S. Government. 
We are grateful for assistance from Google, Amazon, and Intel.}

\thanks{K. Lee was with the Brain \& Cog. Sci. Dept., Massachusetts Institute of Technology, Cambridge, MA 02139 USA, and is now with the Neuroscience Institute, Princeton University, Princeton, NJ 08544 USA (e-mail: kisuk@princeton.edu).}
\thanks{R. Lu is with the Neuroscience Institute, Princeton University, Princeton, NJ 08544 USA (e-mail: ranl@princeton.edu).}
\thanks{K. Luther is with the Dept. of Physics, Princeton University, Princeton, NJ 08544 USA (e-mail: kluther@princeton.edu).} \thanks{H. Sebastian Seung is with the Dept. of Computer Science and the  Neuroscience Institute, Princeton University, Princeton, NJ 08544 USA. (e-mail: sseung@princeton.edu).}
\thanks{KL, RL, and HSS disclose financial interests in Zetta AI LLC.}
}

\maketitle

\begin{abstract}
We show dense voxel embeddings learned via deep metric learning can be employed to produce a highly accurate segmentation of neurons from 3D electron microscopy images. A ``metric graph'' on a set of edges between voxels is constructed from the dense voxel embeddings generated by a convolutional network. Partitioning the metric graph with long-range edges as repulsive constraints yields an initial segmentation with high precision, with substantial accuracy gain for very thin objects. The convolutional embedding net is reused without any modification to agglomerate the systematic splits caused by complex ``self-contact'' motifs. Our proposed method achieves state-of-the-art accuracy on the challenging problem of 3D neuron reconstruction from the brain images acquired by serial section electron microscopy. Our alternative, object-centered representation could be more generally useful for other computational tasks in automated neural circuit reconstruction.
\end{abstract}


\section{Introduction}
\label{sec:intro}

\IEEEPARstart{N}{euronal} connectivity can be reconstructed from a 3D electron microscopy (EM) image of a brain volume \cite{kornfeld2018progress, lee2019convolutional}. A challenging and important subproblem is the segmentation of the image into neurons. One state-of-the-art approach applies a convolutional network to detect neuronal boundaries \cite{zeng2017deepem3d,beier2017multicut,lee2017superhuman,funke2019large}, which are postprocessed to yield a segmentation. Impressive accuracy has been obtained via this approach, and it has proven hard to beat. For example, the boundary detection net of \cite{lee2017superhuman} has remained at the top of the leaderboard of the SNEMI3D neuron segmentation challenge\footnote{\url{http://brainiac2.mit.edu/SNEMI3D/home}} for the past three years. All the leading entries of the CREMI challenge\footnote{\url{https://cremi.org/}} are also boundary detection nets \cite{funke2019large,bailoni2019generalized}.

Alternatives to boundary detection have been proposed. For example, one can train a convolutional net to iteratively extend one object at a time, as in flood-filling nets (FFNs, \cite{januszewski2016flood,januszewski2018high}) and MaskExtend~\cite{meirovitch2016multipass}. Cross-classification clustering~\cite{meirovitch2019cross} introduces a multi-object tracking technique based on convolutional/recurrent nets. Among such alternative approaches, FFNs have successfully been employed to densely reconstruct neurons from several real world datasets \cite{januszewski2018high,li2019automated,scheffer2020connectome}, and follow very closely the boundary detection net of \cite{lee2017superhuman} on the SNEMI3D leaderboard. Nevertheless, none of them have so far displaced boundary detection nets from the top of the SNEMI3D and CREMI leaderboards.

% Nevertheless such alternative approaches have not displaced boundary detection nets from the top of the SNEMI3D and CREMI leaderboards. 

This paper will show that dense voxel embeddings from deep metric learning can be segmented to significantly outperform the state-of-the-art strong boundary detection net of \cite{lee2017superhuman}, the same architecture that leads the SNEMI3D challenge. Convolutional nets are trained to assign similar embedding vectors to voxel pairs within the same objects and well-separated vectors to voxels from different objects \cite{fathi2017semantic,brabandere2017semantic}. This general approach is by now well-known, but has most often been applied to segment images that contain only a few objects, or a few well-separated instances of each object. Brain images from serial section EM, in contrast, contain many densely intertwined branches of neurons.

\begin{figure*}[!h]
\begin{center}
\includegraphics[width=0.9\textwidth]{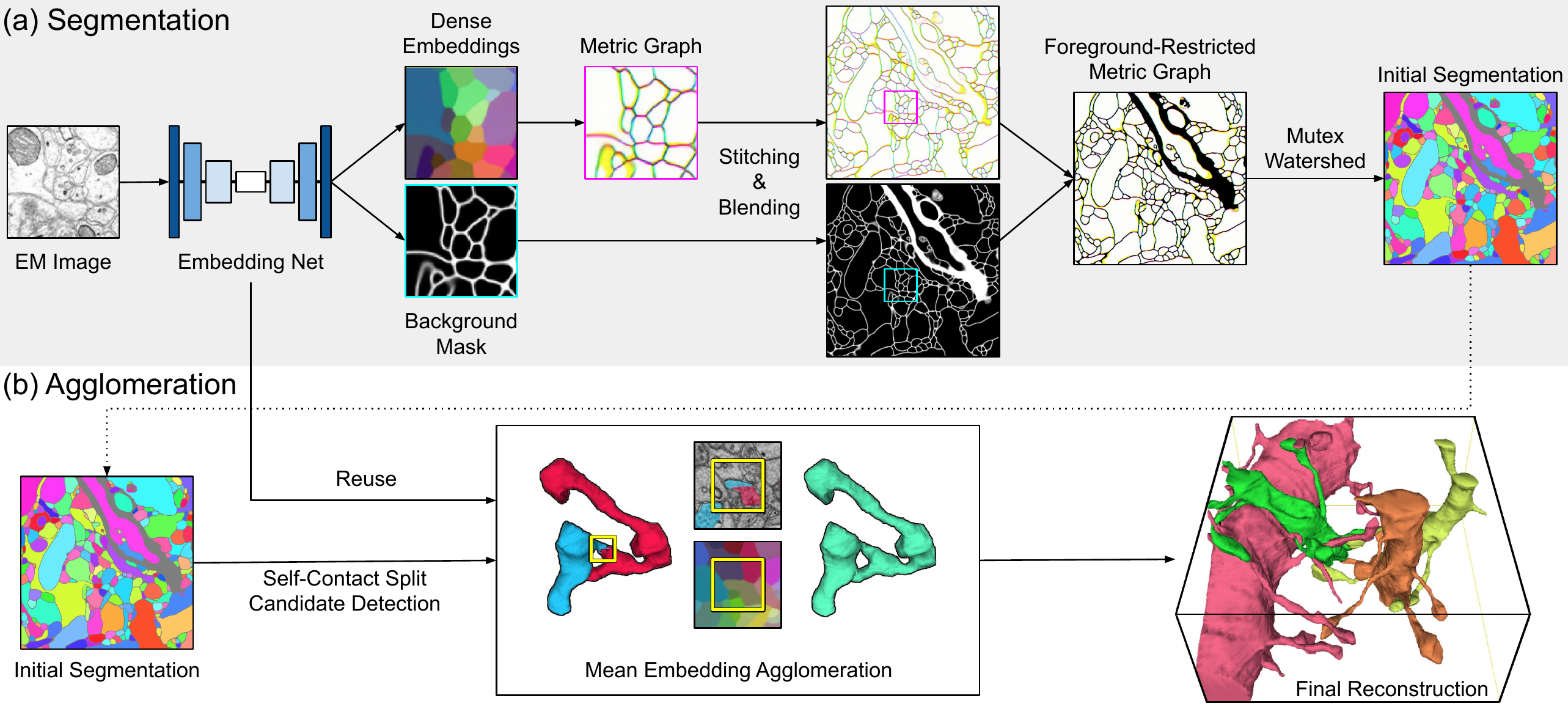}
\caption{Overview of the proposed method. Dense voxel embeddings generated by a convolutional net (Sec.~\ref{sec:loss_function}) are employed at two subtasks for 3D neuron reconstruction: (a) neuron segmentation via \emph{metric graph} \cite{luther2019learning} (Sec.~\ref{sec:metric_graph}) and (b) agglomeration based on mean embeddings (Sec.~\ref{sec:agglomeration}). All graphics shown here (images and 3D renderings) are drawn from real data. Although depicted in 2D for clarity, each 2D image represents a 3D volume. To visualize embeddings, we used PCA to project the 24-dimensional embedding space onto the three-dimensional RGB color space. For brevity, we only visualize nearest neighbor affinities on the metric graph by mapping $x$, $y$, and $z$-affinity to RGB, respectively.} 
\label{fig:overview}
\end{center}
\end{figure*}

Our postprocessing of the voxel embeddings from deep metric learning contains several key elements. First, affinities between voxel pairs are computed from embeddings generated by a convolutional net operating on 3D patches of limited size. The affinities from overlapping patches are stitched and blended to cover the entire image volume (Fig. \ref{fig:overview}). Following \cite{luther2019learning}, we use the term ``metric graph'' to refer to an affinity graph generated by deep metric learning. Second, the metric graph is segmented by the recently proposed Mutex Watershed \cite{wolf2018mutex}, which exploits long-range as well as short-range affinities. Third, the segmentation is improved by agglomerating pairs of objects with similar mean embedding vectors. Combining these three elements significantly outperforms the state-of-the-art boundary detection net of \cite{lee2017superhuman}.

The rationale for the above elements is as follows. The restriction to patches of limited size is required by the memory constraint of the GPU. It also has the effect of making the problem easier for the net by limiting the number, size, and complexity of objects contained in the patch. 

Naively, one might attempt to stitch and blend embeddings rather than affinities, but this is problematic if the embeddings from two patches are inconsistent in the overlap region.

A great deal of previous work has relied solely on nearest neighbor affinities for segmentation \cite{lee2017superhuman, funke2019large}. For example, even when long-range affinities were predicted during training, segmentation at test time was based only on nearest neighbor affinities \cite{lee2017superhuman}. A previous application of deep metric learning to 2D neuron segmentation used only nearest neighbor affinities at test time \cite{luther2019learning}, but we found that this did not yield state-of-the-art segmentation accuracy in 3D. By employing the Mutex Watershed \cite{wolf2018mutex} we were able to leverage long-range affinities to improve segmentation accuracy. 

The Mutex Watershed and mean affinity agglomeration \cite{lee2017superhuman, funke2019large} suffer from a failure mode caused by self-contact. For example, a dendritic spine occasionally bends back and contacts the shaft or an adjacent dendritic spine. Such self-contact is problematic because the existence of a boundary there is local evidence that the two contacting parts of the neuron should lie in two different segments. This local evidence may cause the segmentation algorithm to make a split error. We show that many of the self-contact errors can be corrected by agglomerating segments with similar mean embedding vectors. 

Our work has three novelties. First, we apply deep metric learning to 3D neuron reconstruction (Sec. \ref{sec:loss_function}), building on previous work in 2D \cite{luther2019learning}. Second, we combine the Mutex Watershed with a convolutional net trained to generate voxel embeddings (Sec. \ref{sec:metric_graph}), whereas it was originally combined with a net trained to directly predict affinities \cite{wolf2018mutex}. Third, we recognize the self-contact failure mode, and propose a method of correction using \emph{mean embedding agglomeration} (Sec. \ref{sec:agglomeration}).

Remarkably, we find that deep metric learning outperforms boundary detection by a wide margin, when quantified by variation of information \cite{nunez-iglesias2013machine} or number of errors. Qualitatively, the accuracy gains come from avoiding split errors for very thin objects and for objects with self-contact.

\section{Related Work}
\label{sec:related_work}

\subsection{Boundaries vs. Objects}
State-of-the-art methods for serial section EM images (reviewed in \cite{lee2019convolutional}) employ convolutional nets to detect neuronal boundaries~\cite{lee2017superhuman,funke2019large}, for which \emph{affinity graphs} \cite{turaga2010convolutional} have been widely adopted as underlying representations. 

One could argue that the task of detecting boundaries does not force the net to learn about objects, and that more object-centered representations could be critical for further improvements in accuracy. FFNs \cite{januszewski2016flood,januszewski2018high} and MaskExtend~\cite{meirovitch2016multipass} extend a single object at a time. Cross-classification clustering~\cite{meirovitch2019cross} extends multiple objects simultaneously. Deep metric learning can also be seen as object-centered, but all objects in the image are processed simultaneously.

\subsection{Postprocessing of Affinity Graphs}
Affinity graphs are partitioned to produce an oversegmentation into supervoxels, typically using watershed-type algorithms~\cite{zlateski2015image}. Supervoxels can be greedily agglomerated based on simple statistics such as mean affinity~\cite{lee2017superhuman} or percentiles of binned affinity \cite{funke2019large}. Supervoxels can also be agglomerated by optimizing a global objective for graph partitioning, as in the Multicut problem \cite{beier2017multicut}.

Recently, \cite{wolf2018mutex} has proposed the Mutex Watershed, a greedy algorithm for partitioning signed graphs with both attractive/repulsive edges. This algorithm obviates the need for explicit seeds and tunable thresholds for partitioning, and directly produces a deterministic segmentation, bypassing supervoxel generation. The Mutex Watershed has more recently been integrated into GASP~\cite{bailoni2019generalized}, a generalized framework for signed graph partitioning, and has also been extended to incorporate semantics for joint graph partitioning and labeling~\cite{wolf2019semantic}.

\subsection{Refinement of Neuron Segmentation}
Biological domain knowledge such as geometric~\cite{matejek2019biologically} or semantic \cite{krasowski2018neuron,pape2019leveraging} properties of neurons can be leveraged for agglomeration of supervoxels. Further refinement of segmentation is possible with error detection and correction based on convolutional nets \cite{zung2017error,dmitriev2018efficient}. Unsupervised embeddings of neuronal morphology \cite{schubert2019learning} and supervised classification of neuronal compartments \cite{li2020neuronal} have recently been proposed to detect and correct certain types of merge errors.

% Reference \cite{schubert2019learning} propose to learn unsupervised embeddings of neuronal morphology, which can be used to detect glia and correct glia-neuron mergers. 
% Recently, \cite{li2020neuronal} has proposed supervised classification of neuronal compartments, which can be used to detect and correct certain types of merge errors between distinct neuronal compartments.

We propose \emph{mean embedding agglomeration} (Sec. \ref{sec:agglomeration}) to refine segmentation by agglomerating pairs of objects with similar mean embedding vectors. Independently of our work,\footnote{Our work has been available in preprint form \cite{lee2019learning}.} \cite{han2020occuseg} has proposed to use mean embeddings as one of the several features for agglomerating supervoxels in the 3D instance segmentation of indoor scene. Unlike their method, mean embeddings are our sole feature for agglomeration, and our computation of mean embeddings is restricted to a focused region at the contact between objects, thereby increasing agglomeration accuracy.

\subsection{Deep Metric Learning for 2D Neuron Segmentation}
Our work builds on a previous application of deep metric learning to 2D neuron segmentation \cite{luther2019learning}. Using the loss function of \cite{brabandere2017semantic}, they train a 2D convolutional net to generate dense embeddings, from which nearest neighbor affinities between pixels are computed. The resulting affinity graph, called ``metric graph,'' \cite{luther2019learning} is then partitioned with connected components to yield a segmentation.

% [describe metric graph and other ideas that are used here]

We extend their method to 3D neuron reconstruction, which requires fundamental changes in the way we generate and exploit the dense embeddings. While they trained a 2D convolutional net to generate dense embeddings on the entire 2D image slice, the extent of 3D dense embeddings is severely limited by the memory constraint of the GPU during both training and inference. 

Moreover, a 2D image \emph{slice} contains neuronal \emph{cross sections} of relatively simple shape, which are often locally confined within a small space and thus well-isolated. In contrast, a 3D image \emph{volume} contains intertwined branches of neurons with complex morphology, which extend from one end of the volume to another most of the time. As a consequence, increasing the input size in 2D would not necessarily increase the size and complexity of contained objects, whereas doing so in 3D would obviously increase the both. This poses a fundamental challenge to learning dense embeddings in 3D.
\subsection{Dense Embeddings via Deep Metric Learning}

The concept of learning \emph{dense embeddings} with convolutional nets first appeared in \cite{harley2015learning}. The core idea is to densely map each pixel in the image to a vector in the embedding space, such that the learned embeddings are useful for downstream tasks. The first applications of dense embeddings include semantic segmentation \cite{harley2015learning} and multi-person pose estimation \cite{newell2017associative}. 

Dense embeddings soon started to gain popularity in the problem of semantic instance segmentation \cite{fathi2017semantic,brabandere2017semantic}. In this approach, the loss function encourages the embedding vectors to form \emph{discriminative} clusters for each object instance, and the clustered embeddings are directly utilized to produce an instance segmentation. While \cite{fathi2017semantic} implements an embedding loss by randomly sampling image pixel pairs, the loss function proposed by \cite{brabandere2017semantic} is centered around mean embeddings and their interactions. The latter has extensively been applied to a wide variety of computer vision problems \cite{pham2019jsis3d,lahoud2019instance,halupka2019deep,tian2019learning,han2020occuseg}. 

Instead of the Euclidean embedding of \cite{fathi2017semantic,brabandere2017semantic}, the hyperspherical embedding based on cosine similarity has also been proposed in several applications \cite{kong2018recurrent,xie2019object,payer2018instance,payer2019segmenting,chen2019instance}. 
Notably, \cite{payer2018instance,payer2019segmenting,chen2019instance} show that locally exerting inter-cluster repulsive forces only to spatially neighboring instances is effective at cell segmentation and tracking from light microscopy images, which contain sparsely dispersed and well-isolated cells as opposed to the densely packed and intertwined neurons in EM images.

Our proposed method bears similarities with the method of \cite{Konopczynski2018instance} for instance segmentation of fibers in polymer material. Using the loss of \cite{brabandere2017semantic}, they train a 3D convolutional net to learn dense voxel embeddings for local patches of 3D CT scans. The main difference is in postprocessing; they directly segment each patch-wise embeddings and then iteratively stitch the segmentations between neighboring patches with overlap. Their simple stitching strategy based on the distance between segmented fibers was prone to merge errors \cite{Konopczynski2018instance}.

%-------------------------------------------------------------------------
\section{Methods}
\label{sec:methods}

\subsection{Loss Function}
\label{sec:loss_function}

We use the loss function of \cite{brabandere2017semantic}, which was applied by \cite{luther2019learning} to 2D neuron segmentation. They refer to it as ``means-based loss,'' because the mean embeddings of distinct objects act as cluster centers. The ``internal'' term of \eqref{eq:int} pulls embeddings toward the respective cluster centers, the ``external'' term of \eqref{eq:ext}  pushes distinct cluster centers apart from each other, and the regularization term of \eqref{eq:reg} prevents all cluster centers from deviating too far from the origin:
\begin{gather}
    \mathcal{L}_\text{int} = \frac{1}{C}\sum_{c=1}^{C}\frac{1}{N_c}\sum_{i=1}^{N_c} \lVert \mu_c - x_i \rVert^2, \label{eq:int}
\end{gather}
\begin{gather}
    \mathcal{L}_\text{ext} = \frac{1}{C(C-1)}  \mathop{\sum_{c_A=1}^{C}\sum_{c_B=1}^{C}}_{c_A \neq c_B} \max(2\delta_d - \lVert \mu_{c_A} - \mu_{c_B} \rVert, 0)^2, \label{eq:ext}
\end{gather}
\begin{gather}
    \mathcal{L}_\text{reg} = \frac{1}{C}\sum_{c=1}^{C}\lVert \mu_c \rVert. \label{eq:reg}
\end{gather}
Here $C$ is the number of ground truth objects, $N_c$ is the number of voxels in object $c$, $\mu_c$ is the mean embedding for object $c$, $x_i$ is an embedding for voxel $i$, $\lVert \cdot \rVert$ is the L1 norm, $\delta_d$ is the margin for the external loss term \eqref{eq:ext}. We choose $\delta_d = 1.5$ following \cite{brabandere2017semantic,luther2019learning}. Note that only those voxels that belong to the ground truth objects (i.e., foreground objects) are taken into account in \eqref{eq:int}--\eqref{eq:reg}. The embeddings for background voxels are not included in the loss. 

The embedding loss $\mathcal{L}_\text{embedding}$ is a weighted sum of the three terms,
$
    \mathcal{L}_\text{embedding} = \alpha \mathcal{L}_\text{int} + \beta \mathcal{L}_\text{ext} + \gamma \mathcal{L}_\text{reg}$,
%    \label{eq:embedding}
where $\alpha = \beta = 1$ and $\gamma = 0.001$ as in \cite{luther2019learning}.

Besides the embedding loss, we additionally predict a voxel-wise background mask (Fig.~\ref{fig:overview}) using the standard binary cross-entropy loss. The total loss is a sum of the embedding and background losses,
$
    \mathcal{L}_\text{total} = \mathcal{L}_\text{embedding} + \mathcal{L}_\text{background}$.

It is challenging for a convolutional net to generate embedding vectors that are uniform across neurons, because of their complex and extended morphologies. For achieving approximate uniformity in 3D, we found the means-based loss of \cite{brabandere2017semantic} to be superior to other proposed loss functions for deep metric learning, much as \cite{luther2019learning} found in 2D.

Sometimes parts of the same neuron may seem to be distinct objects when they are restricted to a local patch of limited context. 
For achieving better generalization performance, \cite{luther2019learning} found it critical to locally recompute connected components of the ground truth objects in each training patch, and then remove from \eqref{eq:ext} the pairwise interactions between the locally split object parts. We confirmed their finding in our preliminary experiment, and therefore used the same trick.

\subsection{Segmentation with Metric Graph}
\label{sec:metric_graph}

We define an affinity $a_{ij} \in [0,1]$ between a pair of voxels $i$ and $j$ by
\begin{equation}
    a_{ij} = \max \left(\frac{2 \delta_d - \lVert x_i - x_j \rVert}{2 \delta_d}, 0\right)^2,
    \label{eq:affinity}
\end{equation}
where $x_i$ and $x_j$ are embeddings for the voxel $i$ and $j$, respectively. Note that this definition is directly derived from the term $\max \left(2 \delta_d - \lVert \mu_{C_A} - \mu_{C_B} \rVert, 0\right)^2$ in \eqref{eq:ext}; we simply normalize it to $[0,1]$ after replacing the mean embeddings $\mu_{C_A}, \mu_{C_B}$ with the voxel embeddings $x_i, x_j$. The affinity $a_{ij}$ approaches 1 as $x_i$ and $x_j$ get closer in the embedding space, whereas $a_{ij}$ approaches 0 as $x_i$ and $x_j$ become more distant. 

Once the embedding net generates dense voxel embeddings for a local image patch, we can construct a patch-wise \emph{metric graph}~\cite{luther2019learning} whose nodes are voxels and edge weights are metric-derived affinities between voxel pairs. 

Reference \cite{luther2019learning} explored only the simplest possible form of metric graph and postprocessing, i.e., nearest neighbor metric graph partitioned by connected components clustering. However, we found in our 3D experiments that nearest neighbor affinities are occasionally noisy, primarily due to the noisy embeddings for the background voxels, which were excluded from the embedding loss and thus did not receive explicit training signals (Supp. Fig.~\ref{fig:background}). 

To address this issue, and to push the segmentation accuracy further, we augmented our metric graph and its postprocessing by (a) incorporating long-range affinities into the metric graph and exploiting them as repulsive constraints during graph partitioning, and (b) removing noisy affinities from the metric graph using a voxel-wise background mask predicted by the embedding net (Fig.~\ref{fig:overview}). 

To mask out noisy affinities, we first defined background voxels by thresholding the predicted real-valued background mask with $\theta_\text{mask}$ (empirically chosen on the validation set), and then removed the background voxels (nodes) along with every incident affinities (edges) from the metric graph.

The resulting ``foreground-restricted'' metric graph (Fig. \ref{fig:overview}) was used as input to the Mutex Watershed  \cite{wolf2018mutex}, a recently proposed algorithm for partitioning a graph with both attractive/repulsive edges. They trained a 2D convolutional net to directly predict affinities on a small predefined set of short and long-range edges, thus only a \emph{fixed} graph could be generated. We have no such limitation; affinities on \emph{any} edge can be dynamically computed from the dense voxel embeddings, being much more flexible.

\subsection{Mean Embedding Agglomeration}
\label{sec:agglomeration}

\begin{figure}[t]
\begin{center}
\includegraphics[width=1.0\columnwidth]{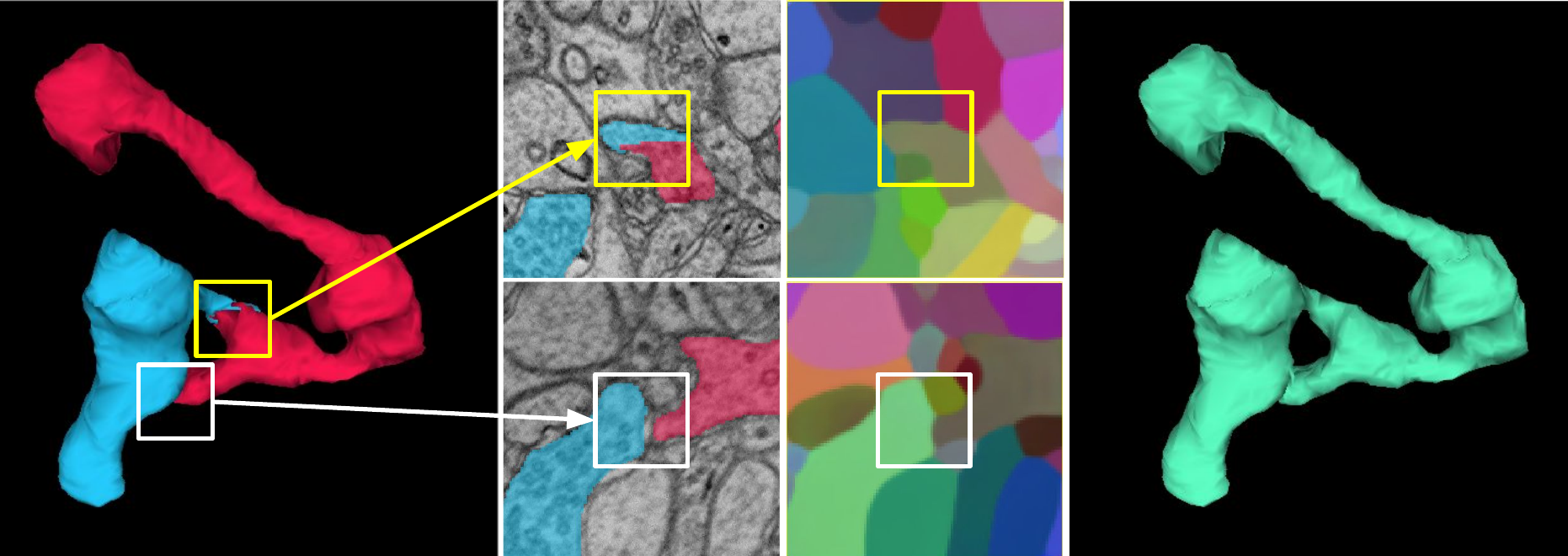}
\caption{Mean embedding agglomeration (Sec.~\ref{sec:agglomeration}). Shown here is a real example of self-contacting axon from the validation set. {\bf Left:} a false split error (yellow box) in the Mutex Watershed~\cite{wolf2018mutex} segmentation caused by a self-contact (white box). {\bf Middle top:} PCA visualization reveals homogeneous embeddings across the false split. {\bf Middle bottom:} each side of the self-contact receives distinct embedding vectors (white box). Note that the both sides appear to be distinct objects in the limited local context. {\bf Right:} mean embedding agglomeration correctly heals the false split.} \label{fig:agglomeration}
\end{center}
\end{figure}

We observed that both the Mutex Watershed~\cite{wolf2018mutex} for our proposed method and mean affinity agglomeration \cite{lee2017superhuman} for the baseline (Sec.~\ref{sec:baseline}) make  systematic split errors on objects with self-contact (Fig.~\ref{fig:agglomeration}). These greedy clustering/agglomeration algorithms suppress localized mistakes in the input graph by averaging out noisy affinities (mean affinity agglomeration) or by putting long-range repulsive constraints as a safeguard in locations with more certainty (the Mutex Watershed). However, they often fail to reconcile the local evidence for disconnectivity at self-contacts (e.g., white box in Fig.~\ref{fig:agglomeration}) with the evidence for connectivity elsewhere (e.g., yellow box in Fig.~\ref{fig:agglomeration}), if the former precedes the latter in greedy decision making.

To address this systematic failure mode, we detected the candidates for self-contact split errors based on a simple and intuitive heuristic. Specifically, we first constructed a region adjacency graph (RAG) from the initial segmentation, where nodes are segments and edges represent spatial adjacency, i.e., whether two segments are contacting each other. Then the foreground-restricted metric graph was reused to compute an agglomeration score $S$ of each individual contact between adjacent segments. To compute $S$, we averaged nearest neighbor affinities between interfacing voxels at each contact.

We selected a pair of adjacent segments as candidate if (a) they have multiple contacts, and (b) the highest $S$ among the multiple contacts is above a certain threshold $\theta_\text{self-contact}$ (chosen empirically on the training and validation sets). Intuitively, self-contact splits should have at least two contacts, false split and self-contact (Fig.~\ref{fig:agglomeration}), and the false split is likely to have high agglomeration score. 

Each candidate is a triplet $(s_1, s_2, (x, y, z))$, where $s_1$ and $s_2$ are the pair of candidate segments and $(x,y,z)$ is the centroid coordinate of the contact with the highest $S$. Given a list of candidates detected by the heuristic, we reused the embedding net without any modification to make an agglomeration decision (Fig.~\ref{fig:agglomeration}). Specifically, we generated dense voxel embeddings on each local image patch centered on $(x,y,z)$, and then computed mean embeddings $\mu_{s_1}$ and $\mu_{s_2}$ of the restrictions of $s_1$ and $s_2$ to a central ``focal'' widow of size $p_x \times p_y \times p_z$ (e.g., yellow box in Fig.~\ref{fig:agglomeration}). The two candidate segments were agglomerated if the L1 distance $d = \lVert \mu_{s_1} - \mu_{s_2} \rVert$ is below a predetermined threshold $\theta_d$. The parameters $p_x$, $p_y$, $p_z$, and $\theta_d$ were determined empirically on the training and validation sets.

\subsection{Establishing a Strong Baseline}
\label{sec:baseline}

As a strong baseline, we reproduced the state-of-the-art result of \cite{lee2017superhuman}, which is currently leading the SNEMI3D benchmark challenge. They train a 3D variant of U-Net \cite{ronneberger2015unet} to directly predict nearest neighbor affinities as a primary training target \cite{turaga2010convolutional}, and long-range affinities as an auxiliary training target. The predicted nearest neighbor affinities are then partitioned with a variant of watershed algorithm \cite{zlateski2015image} to produce an initial oversegmentation. Although the costly 16$\times$ test-time augmentation \cite{zeng2017deepem3d, lee2017superhuman} produced the best result (the top entry of the SNEMI3D leaderboard), mean affinity agglomeration without test-time augmentation was shown to be competitive while being efficient \cite{lee2017superhuman}. Therefore, we included both postprocessing methods in our strong baseline results.

Additionally, we performed a preliminary experiment with baseline models to test the following hypothesis: can we obtain segmentation with comparable accuracy from the images with lower resolution? This is an important question because connectomics is facing an imminent challenge of scaling up to petascale data~\cite{lichtman2014}. We performed dual experiments with the original images (voxel resolution: $6 \times 6 \times 29$ nm$^3$) and $2\times$ in-plane-downsampled images (voxel resolution: $12 \times 12 \times 29$ nm$^3$), and found no significant drop in accuracy (Supp. Fig.~\ref{fig:AC3_test_merge_split}). Therefore, we performed our main experiment exclusively using the $2\times$ in-plane-downsampled images.

\begin{table*}[ht!]
\begin{center}
\caption{Test Set Evaluation Result}\label{tab:vi}
\begin{tabular}{|l|l|c|c|c|}
\hline
Method & Postprocessing & $VI\downarrow$ & $VI_\text{split}\downarrow$ & $VI_\text{merge}\downarrow$\\
\hline
\hline
Baseline$^\dagger$ & Watershed~\cite{zlateski2015image} & 0.0798 & 0.0574 & 0.0224 \\
% Baseline$^\dagger$ & Watershed~\cite{zlateski2015image} + Test-Time Aug. ($16\times$) & 0.0547 & 0.0150 & 0.0697 \\
Baseline$\star^{\dagger}$ & Watershed~\cite{zlateski2015image} + Test-Time Augmentation ($16\times$) \cite{zeng2017deepem3d,lee2017superhuman} & 0.0610 & 0.0475 & {\bf 0.0135} \\
Baseline & Watershed~\cite{zlateski2015image} + Mean Affinity Aggl.~\cite{lee2017superhuman} & 0.1049 & 0.0877 & 0.0172 \\
\hline
\hline
Hybrid & Watershed~\cite{zlateski2015image} + Mean Affinity Aggl.~\cite{lee2017superhuman} + Mean Embedding Aggl. (Sec.~\ref{sec:agglomeration}) & 0.0572 & 0.0399 & 0.0173 \\
\hline
\hline
Proposed & Mutex Watershed~\cite{wolf2018mutex} & 0.1188 & 0.1025 & 0.0163 \\
Proposed & Mutex Watershed~\cite{wolf2018mutex} + Mean Embedding Aggl. (Sec.~\ref{sec:agglomeration}) & {\bf 0.0470} & {\bf 0.0276} & 0.0194 \\
\hline
% \multicolumn{5}{p{430pt}}{$\star$ For transparency, we included the best baseline result that arose unexpectedly from a preliminary experiment without the ``slip interpolation'' data augmentation (Sec.~\ref{sec:exp}).}\\
\multicolumn{5}{p{430pt}}{$\dagger$ For these baselines, we chose the best possible operating point by optimizing postprocessing parameters directly on the test set, giving advantage over our proposed method.}\\
\multicolumn{5}{p{430pt}}{$\downarrow$ The lower, the better.}\\
\end{tabular}
\end{center}
\end{table*}

%-------------------------------------------------------------------------
\section{Experimental setup}
\label{sec:exp}

\subsection{Dataset}
\label{sec:dataset}
Since its 2013 launch, the SNEMI3D benchmark challenge has catalyzed remarkable progress in automated neuron reconstruction algorithms \cite{beier2017multicut,zeng2017deepem3d,lee2017superhuman,meirovitch2019cross}. As mentioned earlier, self-contact is a major failure mode for one of the leading SNEMI3D submissions \cite{lee2017superhuman}. We noticed a qualitative difference in neuronal morphology between SNEMI3D's training and test sets. While the test set contains large spiny dendrites with many self-contacting spines, the training set barely contains them. Therefore we decided to create a training set containing more examples of self-contact. 

We used the publicly available AC3/AC4 dataset,\footnote{\url{https://software.rc.fas.harvard.edu/lichtman/vast/AC3AC4Package.zip}} which is a superset of SNEMI3D. AC3 and AC4 are human-labeled subvolumes from the mouse somatosensory cortex dataset of \cite{kasthuri2015saturated}, which was acquired by serial section EM. The sizes of AC3 and AC4 are $1024 \times 1024 \times 256$ and $1024 \times 1024 \times 100$ voxels, respectively, at $6 \times 6 \times 29$ nm$^3$ voxel resolution. We used AC4 and the bottom 116 slices of AC3 for training, the middle 40 slices of AC3 for validation, and the top 100 slices of AC3 for testing.

The SNEMI3D training set is AC4 only, and the SNEMI3D test set is the top 100 slices of AC3. In other words, this paper's training set is the SNEMI3D training set plus some of AC3. This paper's test set is the same as the SNEMI3D test set, except that extra image padding was obtained from the full image stack of \cite{kasthuri2015saturated},\footnote{\url{https://neurodata.io/data/kasthuri15/}} and used at test time only to provide enough image context for preventing systematic drop in accuracy near the dataset edge. 

% Since this paper uses an enlarged training set, we have not submitted the results to the SNEMI3D leaderboard. Instead, 
Since this paper uses an enlarged training set, we are retraining the leading SNEMI3D submission \cite{lee2017superhuman} and using it as a strong baseline for comparison in our main experiment (Sec. \ref{sec:results}). We note that SNEMI3D has not, strictly speaking, been a blind challenge since the publication of AC3/AC4 \cite{kasthuri2015saturated}. Submissions to SNEMI3D are on an honor system. To enable objective comparison with other methods, we have also retrained our proposed method strictly under the SNEMI3D challenge setup and submitted the result to the leaderboard (Supp. Note). Our code and data will be made available at \url{https://github.com/seung-lab/devoem}.

\subsection{Network Architecture}
\label{sec:network}

We used a modified version of the ``Residual Symmetric U-Net'' architecture of \cite{lee2017superhuman}. We used a $128\times128\times20$ voxel input patch for both training and inference, using $2\times$ in-plane-downsampled images. The net produces as output 24-dimensional dense voxel embeddings, as well as a single channel background mask. These outputs are then spatially cropped to $96\times96\times16$ in order to reduce uncertainty near the patch border. To give more expressive power, we linearly scaled the embeddings with a learnable scalar parameter, which was initialized to $0.1$ at the beginning of training. For upsampling, we used the bilinear \emph{resize convolution} \cite{odena2016deconvolution}, i.e., bilinear upsampling followed by a pointwise ($1\times1\times1$) convolution. Further details are illustrated in Supp. Fig.~\ref{fig:embedding_net}.

\subsection{Data Augmentation}
\label{sec:augmentation}
We used the same training data augmentation as in \cite{lee2017superhuman}. This includes flip \& rotation by $90^\circ$, brightness \& contrast perturbation, warping, simulated misalignment, simulated out-of-focus and missing sections.

Additionally, we introduced a novel ``slip interpolation'' that substitutes the slip-type simulated misalignment of \cite{lee2017superhuman} (Supp. Fig.~\ref{fig:slip_interpolation}). Specifically, we simulated the slip-type misalignment only in the input, not in the target. The mismatch between the input and target forces the nets to ignore any slip misalignment in the input and produce smoothly interpolated target prediction.

\subsection{Training Details} 
We performed all experiments with PyTorch. We trained our nets on four NVIDIA Titan X Pascal GPUs using synchronous gradient update. We used the AMSGrad variant \cite{reddi2018} of the Adam optimizer \cite{kingma2015}, with $\alpha = 0.001, \beta_1 = 0.9, \beta_2 = 0.999$, and $\epsilon = 10^{-8}$. We used a single training patch (i.e., minibatch size of 1) for each model replica on GPUs at each gradient step. We trained the baseline and embedding nets for three and five days, respectively, and selected the model checkpoints at the lowest validation error.

\subsection{Inference} 
We used the overlap-blending inference of \cite{lee2017superhuman}. Both the baseline and embedding nets used the most conservative output overlap of $50\%$ in $x$, $y$, and $z$-dimension. The embedding net's output cropping incurred nearly $2\times$ overhead relative to the baseline, which did not use output cropping. This is because the $50\%$ overlap between \emph{cropped outputs} amounts to the overlap higher than $50\%$ between \emph{inputs}, thus increasing the coverage factor for each input voxel.

\subsection{Metric Graph \& Postprocessing} 
To construct the metric graph as input to the Mutex Watershed \cite{wolf2018mutex}, we used three nearest neighbor \emph{attractive} edges and nine long-range \emph{repulsive} edges. Specifically, each \emph{edge} is characterized by an \emph{offset vector}. An \emph{affinity map} for a given edge is constructed by computing a map of metric-derived affinities (using \eqref{eq:affinity} in Sec.~\ref{sec:metric_graph}) between a reference grid of voxels and another grid of voxels shifted by the characteristic offset vector. The list of $(x,y,z)$ offset vectors we used is (-1,0,0), (0,-1,0), (0,0,-1), (0,0,-2), (-5,0,0), (0,-5,0), (-5,-5,0), (-5,5,0), (-5,0,-1), (0,-5,-1), (-5,0,1), and (0,-5,1). These 12 edges (or offset vectors) yield 12 affinity maps that comprise the resulting metric graph (Supp. Fig. \ref{fig:edge_neighborhood_structure}).

% \kisuk{$2\times$ downsampled long-range affinity maps.}

We grid searched the postprocessing parameters strictly on the training and validation sets. Selected parameters were $\theta_\text{mask} = 0.6$, $\theta_\text{self-contact} = 0.25$, $\theta_d = 1.5$, and $p_x \times p_y \times p_z = 32\times32\times5$. 

\subsection{Evaluation}
For quantitative evaluation of segmentation quality, we adopted the widely-used variation of information (VI), \cite{nunez-iglesias2013machine,arganda-carreras2015crowdsourcing},
$
    VI = VI_\text{split} + VI_\text{merge} = H(S|T) + H(T|S)$,    
where $H(\cdot|\cdot)$ is the conditional entropy, $S$ is the segmentation proposal, and $T$ is the ground truth segmentation.

%-------------------------------------------------------------------------
\section{Results}
\label{sec:results}

\subsection{Quantitative Analysis}
\label{sec:quant}

Table \ref{tab:vi} shows the quantitative evaluation of different methods on the test set. To evaluate segmentation quality, we adopted the widely-used variation of information (VI) error metric \cite{nunez-iglesias2013machine}. Metric graph followed by the Mutex Watershed produced an initial segmentation with high precision, which is indicated by the low $VI_\text{merge}$ of 0.0163. However, the initial segmentation largely suffered from the systematic self-contact split errors (Sec.~\ref{sec:mea_effect}), which are reflected in the high $VI_\text{split}$ of 0.1025. 

Remarkably, mean embedding agglomeration healed almost all of these self-contact split errors (Sec.~\ref{sec:mea_effect}), resulting in a precipitous drop in $VI_\text{split}$ from 0.1025 to 0.0276. With this substantial improvement, our proposed method outperformed all baselines in VI. We found that the slight increase in $VI_\text{merge}$ from 0.0163 to 0.0194 after mean embedding agglomeration was caused by the correct agglomeration of a self-contacting glia fragment, which already contained a small merge error. As exemplified by this, VI is sensitive to the size of erroneous objects, and thus tends to hide underlying qualitative difference especially when comparing highly accurate methods. Therefore, we will present rigorous qualitative analyses in the following sections.

% \begin{figure}[t!]
% \begin{center}
% \includegraphics[width=0.85\columnwidth]{figures/thin_parallel_spine.pdf}
% \caption{Our proposed method (bottom row) correctly segments a very difficult thin spine neck that is parallel to the imaging plane.} \label{fig:thin_parallel_spine}
% \end{center}
% \end{figure}

\subsection{Improvements on Very Thin Objects}
\label{sec:thin}

\begin{figure}[t!]
\begin{center}
\includegraphics[width=0.85\columnwidth]{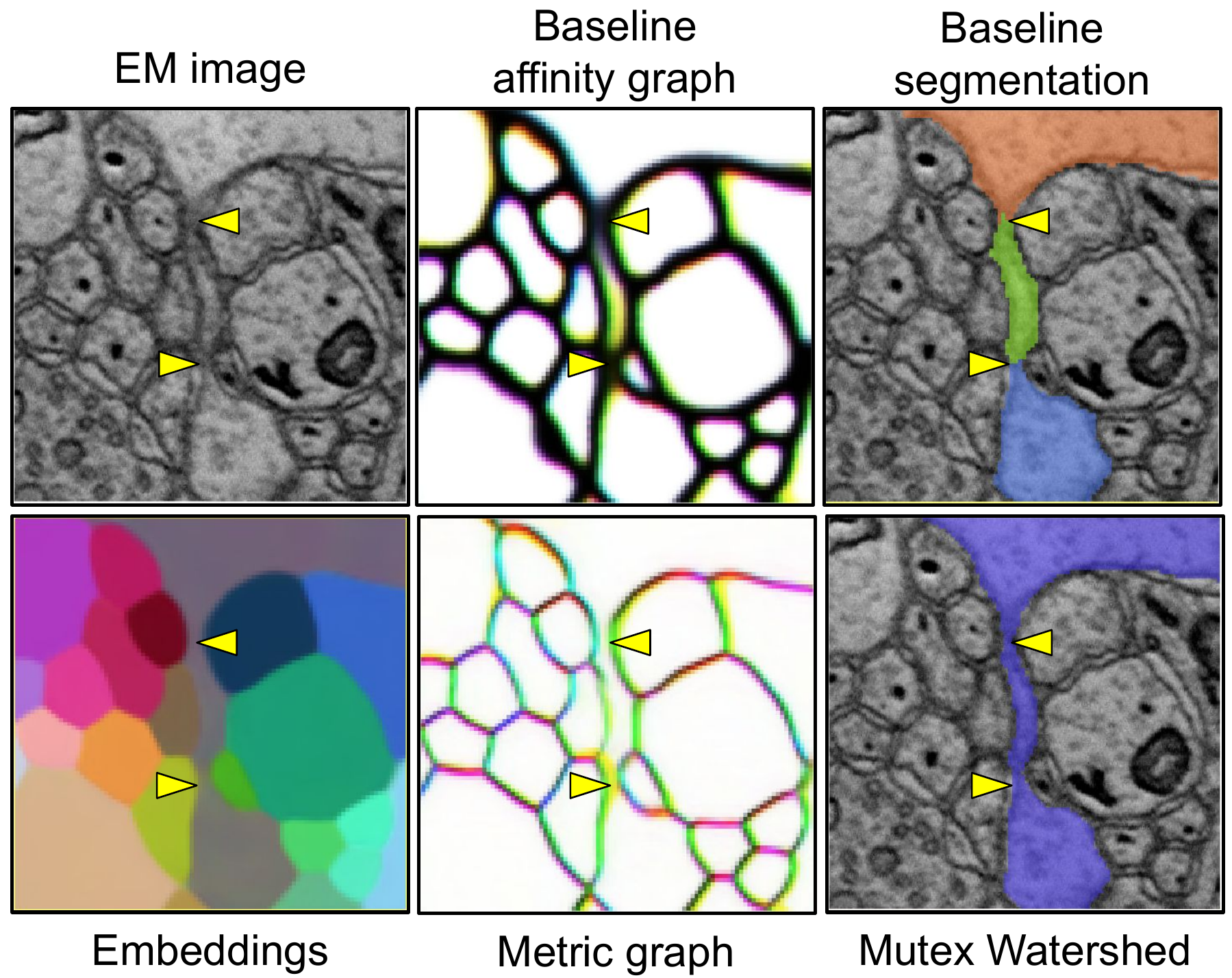}
\caption{Our proposed method (bottom row) correctly segments a very difficult thin spine neck that is parallel to the imaging plane.} \label{fig:thin_parallel_spine}
\end{center}
\end{figure}

\begin{figure}[t!]
\begin{center}
\subfloat[Dendritic spine neck\label{fig:thin_spine}]{
\includegraphics[width=1.0\columnwidth]{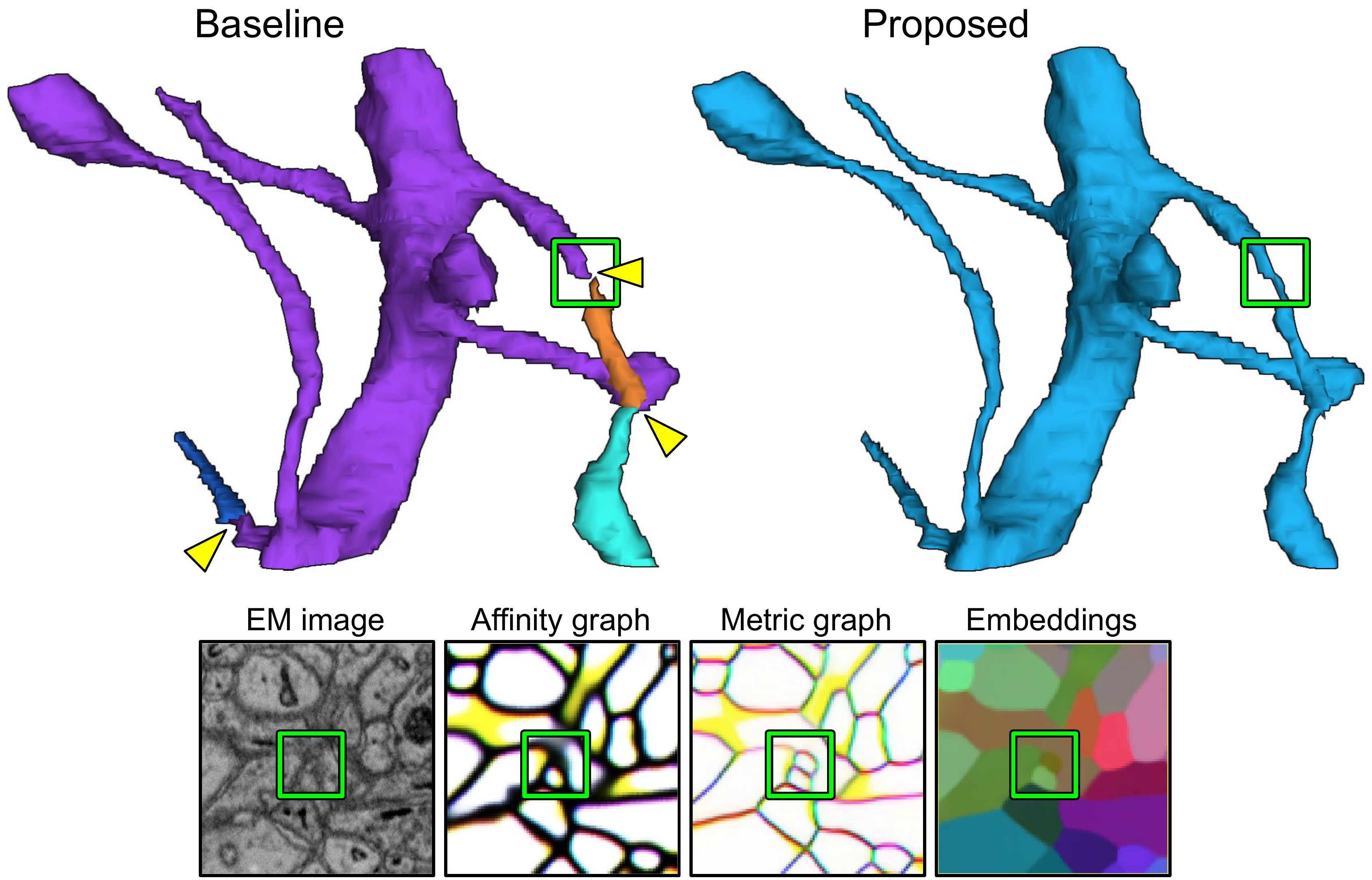}}\\
\subfloat[Axonal branch\label{fig:thin_axon}]{
\includegraphics[width=0.9\columnwidth]{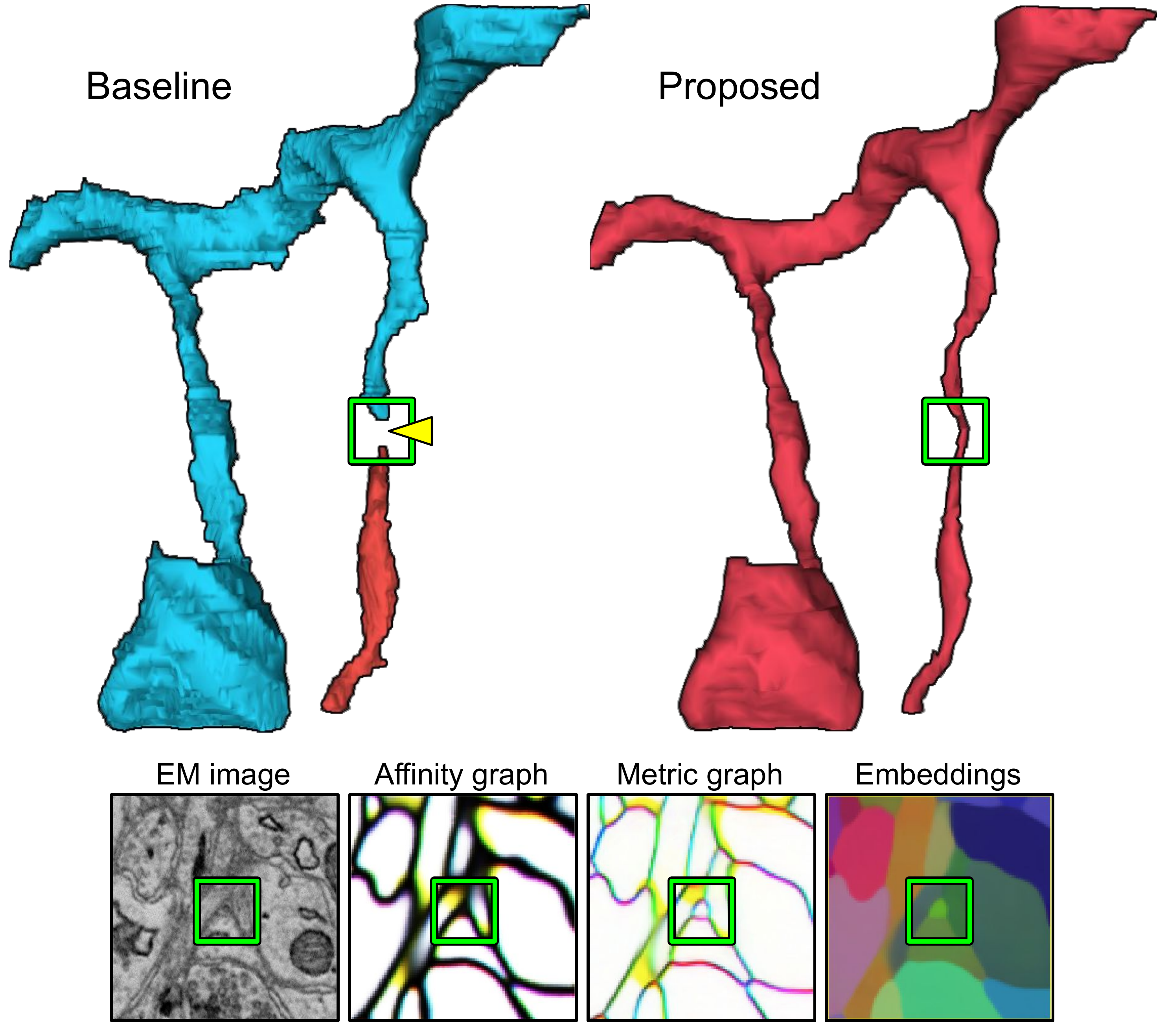}}
\caption{Comparison on the thinnest parts of neurites. Yellow arrowheads indicate split errors.}
\label{fig:thin_neurite}
\end{center}
\end{figure}

Neuronal branches, or neurites, can become very thin, such as in the thinnest part of axons and dendritic spine necks~\cite{helmstaedter2013cellular}. Being capable of tracing such thin neurites is crucial for reconstructing neuronal connectivity. Qualitatively, we observed that our proposed method performs substantially better on very thin objects compared to the baseline. Visualization reveals how the object-centered representation of dense voxel embeddings could outperform the boundary-centered representation of the baseline on the thin objects (Fig.~\ref{fig:thin_parallel_spine}--\ref{fig:thin_neurite}).

Fig.~\ref{fig:thin_parallel_spine} compares segmentation accuracy of the baseline and proposed methods on a very thin spine neck that is parallel to the imaging plane. The thinnest parts (yellow arrowheads in Fig.~\ref{fig:thin_parallel_spine}) are so severely constricted that it is extremely difficult for the boundary-centered baseline to learn and represent the object continuity. In contrast, our embedding net was able to assign similar vectors across the constricted parts, yielding a metric graph that could represent the object continuity adequately (Fig.~\ref{fig:thin_parallel_spine}).

Further examples of extremely thin parts of neurites in different orientations are shown in Fig.~\ref{fig:thin_spine} (oblique spine neck) and Fig.~\ref{fig:thin_axon} (vertical axonal branch). The thinnest parts are again so severely constricted that their cross sections are barely recognizable in the EM images. The boundary-centered baseline totally failed at recognizing the cross sections, whereas our embeddings and metric graph captured them successfully (Fig.~\ref{fig:thin_spine}, \ref{fig:thin_axon}). It is remarkable that learning about object rather than boundary enables the net to virtually ``imagine through'' the input image that lacks much evidence for object continuity.

\begin{figure*}[t!]
\begin{center}
\subfloat[Large spiny dendrite with self-contacting spines\label{fig:self_touching_spines}]{\includegraphics[width=0.8\linewidth]{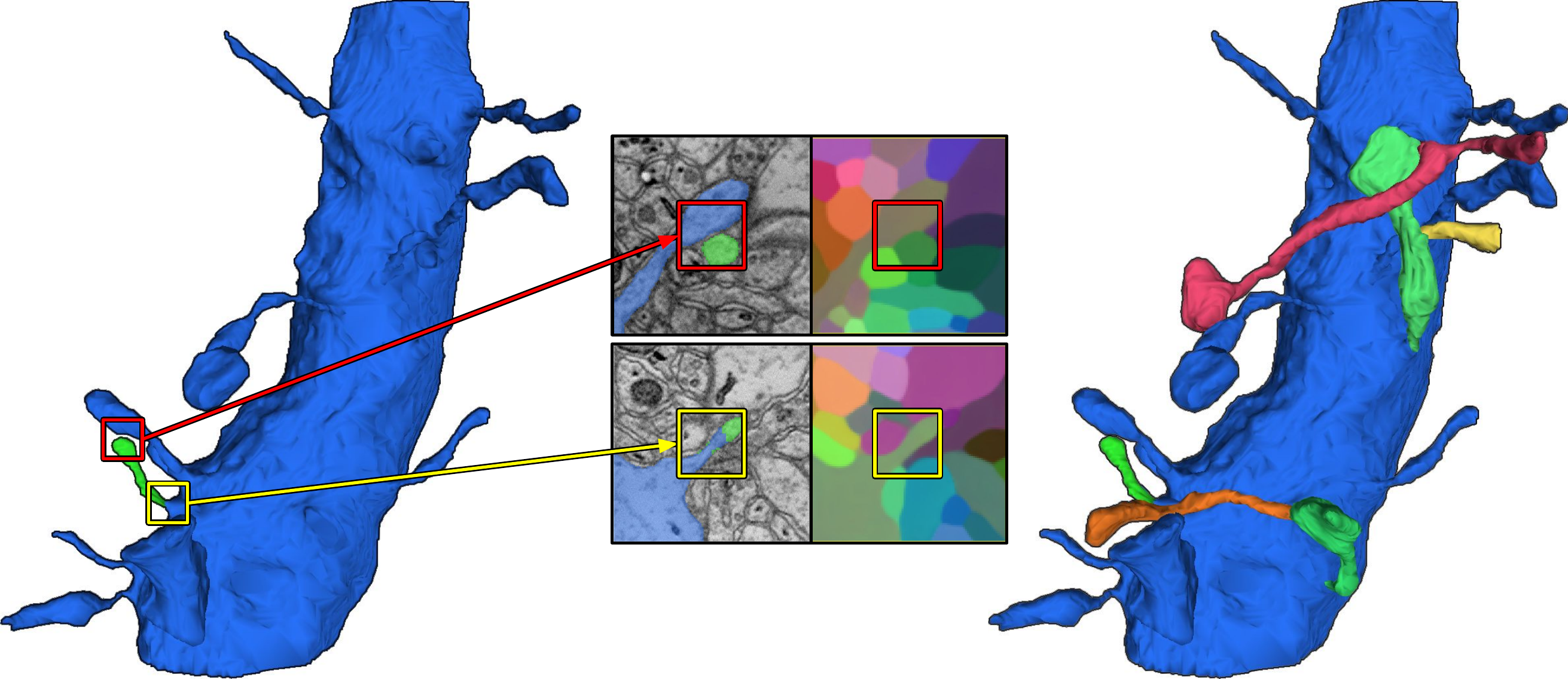}}\\
\subfloat[Complex glia (putative astrocyte) with self-contact\label{fig:self_touching_glia}]{\includegraphics[width=0.68\linewidth]{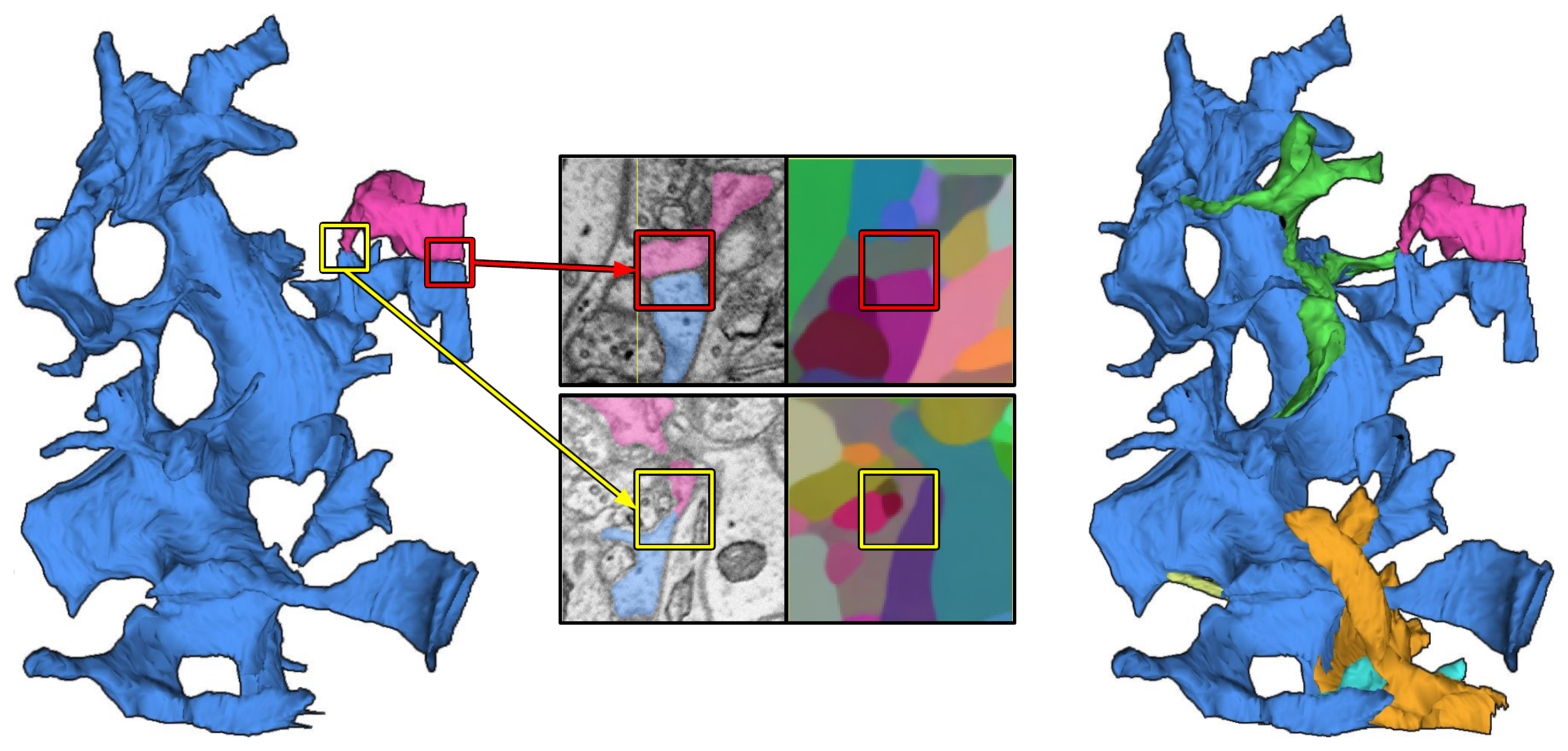}}
\caption{Mean embedding agglomeration successfully corrects most of the self-contact split errors. {\bf Left:} one of the self-contact split errors (top: green spine, bottom: pink glia fragment) is displayed along with the main object (top: dendritic shaft, bottom: glia body). {\bf Middle:} visualization of embeddings at the self-contact (red box) and false split (yellow box). {\bf Right:} final reconstruction after applying mean embedding agglomeration.}
\end{center}
\end{figure*}

% \begin{figure*}[!t]
% \begin{center}
% \includegraphics[width=0.75\linewidth]{figures/self_touching_spines.pdf}
% \caption{Mean embedding agglomeration successfully reattached six out of seven broken self-contacting dendritic spines. Left: one of the broken self-contacting spines in the Mutex Watershed~\cite{wolf2018mutex} segmentation (green) is displayed along with the dendritic shaft (blue). Middle: distinct vectors were assigned to the self-contacting spines (red box), whereas the embeddings were uniform across the false split (yellow box).} \label{fig:self_touching_spines}
% \end{center}
% \end{figure*}

% \begin{figure*}[!t]
% \begin{center}
% \includegraphics[width=0.75\linewidth]{figures/self_touching_glia.pdf}
% \caption{Mean embedding agglomeration successfully reattached all broken self-contacting glia fragments. Left: one of the five broken self-contacting glia fragments in the Mutex Watershed segmentation~\cite{wolf2018mutex} (pink) is displayed along with the main glia body (blue). Middle: distinct vectors were assigned across the self-contact (red box), whereas the embeddings were uniform across the false split (yellow box).} \label{fig:self_touching_glia}
% \end{center}
% \end{figure*}

\subsection{Effectiveness of Mean Embedding Agglomeration}
\label{sec:mea_effect}

As we mentioned earlier, our initial segmentation produced with the Mutex Watershed contained systematic split errors caused by objects with complex self-contact. They were found mostly in large spiny dendrites (Fig.~\ref{fig:self_touching_spines}) and astrocytes, a kind of glia (Fig.~\ref{fig:self_touching_glia}). Since our proposed method is based on dense voxel embeddings for local image patches, their limited context may make a self-contact seem like a boundary between two distinct objects (e.g., red boxes in Fig.~\ref{fig:self_touching_spines}, \ref{fig:self_touching_glia}). Consequently, the Mutex Watershed may put long-range repulsive constraints across the self-contact, making a false split error.

We found that mean embedding agglomeration (Sec. \ref{sec:agglomeration}) is highly effective at agglomerating the self-contact split errors, without introducing new merge errors. Examples from the test set are shown in Fig.~\ref{fig:self_touching_spines}, \ref{fig:self_touching_glia}.

\begin{figure*}[t!]
\begin{center}
\subfloat[Training + validation sets\label{fig:aggl_plot_a}]{\includegraphics[width=\columnwidth]{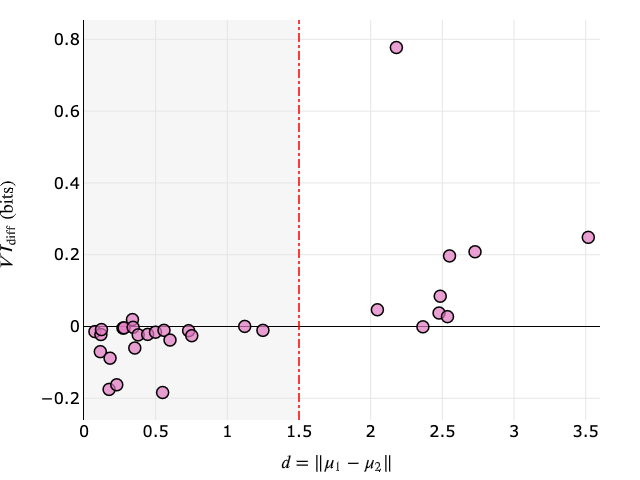}}
\subfloat[Test set\label{fig:aggl_plot_b}]{\includegraphics[width=\columnwidth]{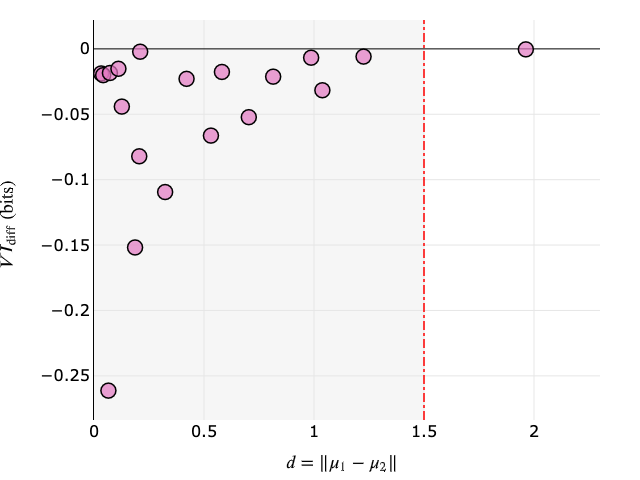}}
\caption{Analysis on individual decisions made by mean embedding agglomeration on (a) training + validation sets and (b) test set. Pinky circles represent individual agglomeration candidates selected by the heuristic described in Sec. \ref{sec:agglomeration}. Vertical axes represent the change in VI after agglomeration, and horizontal axes represent the distance between mean embeddings. See main text for details.} 
\label{fig:aggl_plot}
\end{center}
\end{figure*}

Fig.~\ref{fig:aggl_plot_a} shows an analysis on how we selected the decision boundary $\theta_d$ for mean embedding agglomeration on the training and validation sets. Pink circles represent individual agglomeration candidates selected by the heuristic described in Sec.~\ref{sec:agglomeration}. Horizontal axis represents the L1 distance $d = ||\mu_1 - \mu_2||$ between the mean embeddings of two agglomeration candidates, and vertical axis represents $VI_\text{diff}$, which is the change in VI when the two candidates are \emph{forced} to be agglomerated. Here we computed $VI_\text{diff}$ within each local patch for agglomeration.

We grid searched the agglomeration decision boundary $\theta_d$ on the training and validation sets. Fig.~\ref{fig:aggl_plot_a} shows that the resulting $\theta_d = 1.5$ nicely separates the true positive self-contact splits (left to the decision boundary in Fig.~\ref{fig:aggl_plot_a}) from the false positives (right to the decision boundary in Fig.~\ref{fig:aggl_plot_b}). Although determined empirically, the optimal $\theta_d = 1.5$ turned out to be equivalent to the $\delta_d = 1.5$ of \eqref{eq:ext} in Sec.~\ref{sec:loss_function}, being theoretically reasonable.

% The bottom plot in Fig.~\ref{fig:aggl_plot_a} shows the decomposition of VI into $VI_\text{merge}$ and $VI_\text{split}$. 

Next, we performed a similar \emph{postmortem} analysis on how effective mean embedding agglomeration was on the test set (Fig.~\ref{fig:aggl_plot_b}). Among 19 candidates selected by the heuristic, 18 were agglomerated correctly (true positives, left to the decision boundary in Fig.~\ref{fig:aggl_plot_b}) and one was rejected incorrectly (true negative, right to the decision boundary in Fig.~\ref{fig:aggl_plot_b}). The true negative rejection was caused by a self-contact within the field of view of the embedding net. We found it hard to fix this particular failure mode, which is illustrated with a representative example in Supp. Fig.~\ref{fig:self_touching_within_patch}.

To further demonstrate the effectiveness of mean embedding agglomeration, we additionally applied it to the baseline segmentation that was postprocessed with mean affinity agglomeration, which was also reported to suffer from systematic self-contact split errors~\cite{lee2017superhuman}. Remarkably, mean embedding agglomeration significantly reduced $VI_\text{split}$ without any increase in $VI_\text{merge}$ (``Hybrid'' in Table~\ref{tab:vi}), outperforming all baselines in VI. Nevertheless, this hybrid approach was still slightly inferior to our proposed method (Table~\ref{tab:vi}).

\subsection{Further Evaluation on Extra Test Sets}
\label{sec:further}

\begin{figure}[t!]
\begin{center}
\includegraphics[width=1.0\columnwidth]{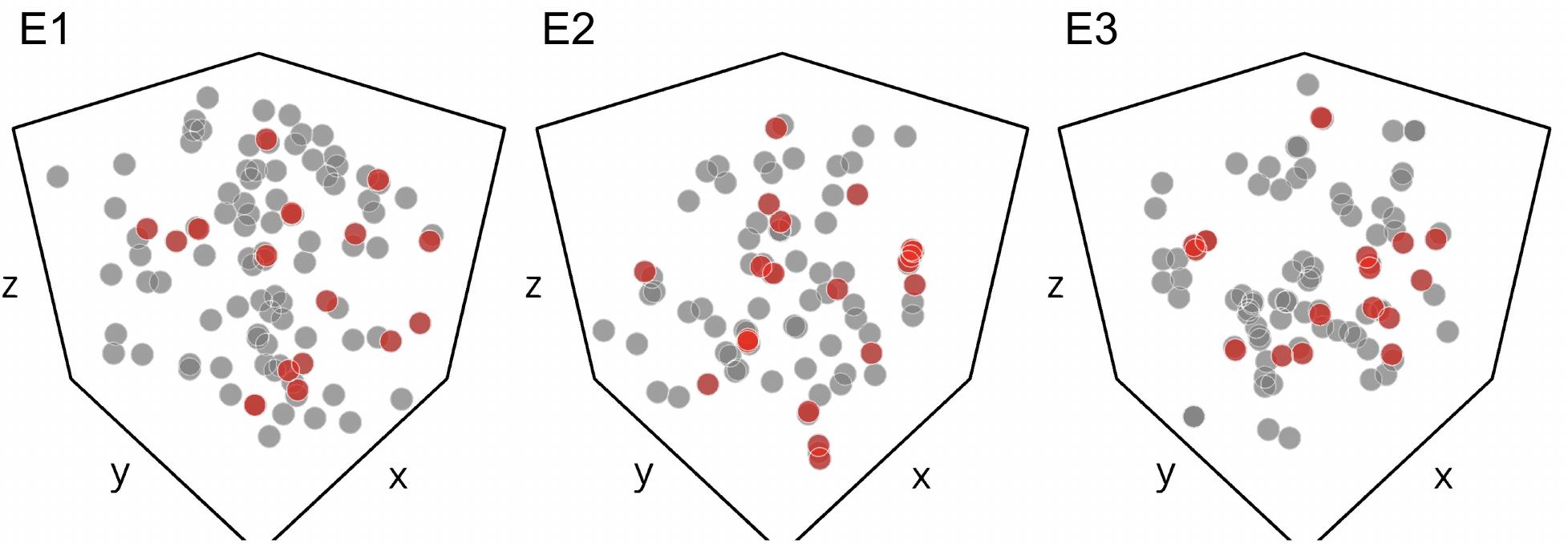}
\caption{Spatial distribution of remaining errors in three extra test volumes E1--E3. Gray circles: remaining errors in the best baseline segmentation. Red circles: remaining errors in the proposed segmentation.} \label{fig:error}
\end{center}
\end{figure}

\begin{table}[t!]
\begin{center}
\caption{Number of Errors in Extra Test Sets}\label{tab:extra}
\begin{tabular}{|c|c|c|c|c|}
\hline
 & E1 & E2 & E3 & Mean $\pm$ S.E.\\
\hline
\hline
Baseline & 87 & 62 & 76 & 75.0 $\pm$ 7.23\\
Proposed & 16 & 20 & 16 & 17.3 $\pm$ 1.33\\
\hline
\end{tabular}
\end{center}
\end{table}

We questioned whether the difference between our proposed method and the baselines could still hold outside the dataset used for our experiments. This is a critical question because we divided the single volume of AC3 into training, validation, and test sets. The outstanding test set performance of our proposed method could have been attributable, at least partly, to statistical similarity between the training, validation, and test sets due to their spatial proximity. 

To address this question, three extra image volumes (namely E1, E2, and E3) of the same size as the test set were obtained from randomly chosen locations in the full dataset of \cite{kasthuri2015saturated} (Supp. Fig.~\ref{fig:E1-E3_location}). We applied both our proposed method and the best baseline (``Baseline$\star$'' in Table~\ref{tab:vi}) on these volumes, and four expert brain image analysts exhaustively examined every single segment to find out remaining errors. 

Fig.~\ref{fig:error} visualizes the spatial distribution of individual merge/split errors from the baseline and proposed method in E1--E3. There were $17.3 \pm 1.33$ and $75.0 \pm 7.23$ remaining errors (mean $\pm$ S.E., $N=3$) from the proposed and baseline method, respectively (Table~\ref{tab:extra}). These results suggest that our proposed method produces a significantly lower number of errors compared to the strong baseline ($p=0.0014$, paired sample t-test), although one should be cautious about the small number of sample volumes ($N = 3$). 

Qualitatively, we observed a similar tendency that our proposed method performs substantially better at very thin objects while mean embedding agglomeration successfully healed most of the self-contact split errors. However, we also found that there remained some uncaught self-contact split errors, mostly due to the heuristic's failure in detecting them (Supp. Fig.~\ref{fig:thin_object_comparison_glia}).

%-------------------------------------------------------------------------
\section{Discussion}
\label{sec:discussion}

% Our proposed method outperforms the strong baseline of \cite{lee2017superhuman}, which is currently leading the SNEMI3D benchmark challenge. In the future, it would be interesting to directly compare our method with FFNs \cite{januszewski2016flood,januszewski2018high}, which closely follow the strong baseline of \cite{lee2017superhuman} on the SNEMI3D leaderboard. 

% Intuitively, one might expect long-range affinities for voxels in different objects to be less noisy than short-range affinities. The reason is that the precise location of a boundary between two objects may be ambiguous, even if the existence of the boundary is not at all in question. 

Learning dense voxel embeddings can be viewed as learning a ``panoptic'' representation of \emph{all} objects in a scene. While being similar in learning an object-centered representation, our approach is contrasted with FFNs \cite{januszewski2016flood,januszewski2018high}, which focus on a \emph{single} object in a scene. This difference may make the two approaches complementary to each other. For instance, \cite{januszewski2018high} employs FFNs for agglomeration as well as for generating an initial oversegmentation. Their FFN-based agglomeration could possibly be complemented with the panoptic representation of dense voxel embeddings. 

In principle, our proposed method is not \emph{linearly} scalable due to the \emph{supra-linear} time complexity of the Mutex Watershed~\cite{wolf2018mutex} algorithm, not to mention the obvious limitation in memory. In practice, such limitations can be overcome by partitioning an entire dataset into tractable subvolumes, which can be independently segmented in parallel and then stitched together to produce a global segmentation. 

% \footnote{Reference \cite{wolf2018mutex} claimed an \emph{empirical} time complexity of $\mathcal{O}(E\log{}E)$, where $E$ is the number of edges in the input graph. They also proved that the time complexity can be $\mathcal{O}(E^2)$ in the worst case.}

Although such a ``block-stitching" approach has been shown to work in practice~\cite{matveev2017multicore,januszewski2018high}, introducing stitching errors may be inevitable especially when using simplistic stitching algorithms, e.g., that are based solely on segmentation overlap between blocks (i.e, subvolumes). To reduce stitching errors while limiting computation overhead, dense voxel embeddings could be selectively employed to resolve only those ambiguous cases in stitching.

We have seen that the self-contact motifs can cause problems in segmentation because they present opposing \emph{local} and \emph{global} connectivity depending on the size of context. We found that they also raise a tricky technical issue when preparing local training patches. As we briefly mentioned in Sec.~\ref{sec:loss_function}, recomputing connected components of the ground truth objects in a local training patch of limited context is an important preprocessing step. With this preprocessing, the self-contacting parts of a neuron should be cleaved into distinct pieces if they are connected outside the local training patch. However, we observed that locally recomputing connected components occasionally fails to cleave such self-contact, depending on whether background voxels properly form a separating ``gap'' at the self-contact (Supp. Fig.~\ref{fig:glial_confusion}). 

Consequently, numerous self-contacts in glia (e.g., see Fig.~\ref{fig:self_touching_glia}) became a significant source of inconsistency that confuses the embedding net during training, resulting in occasional generalization failures specific to glia. Therefore, it would be necessary to carefully remove such inconsistency in the future. Alternatively, targeted detection and special handling of glia could be a remedy to this problem. For instance, the Semantic Mutex Watershed~\cite{wolf2019semantic} combined with glia detection could be effective at preventing merge errors between neurons and glia.

%-------------------------------------------------------------------------
\section{Conclusion}
\label{sec:conclusion}
We have presented a novel application of dense voxel embeddings for 3D neuron reconstruction. We have demonstrated that our proposed method outperforms the state-of-the-art method based on boundary detection~\cite{lee2017superhuman}, and that learning an object-centered representation enables substantial improvements on very thin objects. Future work will include scale-up of the proposed method with block-stitching \cite{matveev2017multicore,januszewski2018high} and large-scale evaluation on real world datasets. Our dense voxel embeddings could be more generally useful for other computational tasks in the connectomics pipeline. For instance, the error detection and correction system of \cite{zung2017error} could be further enhanced with the rich information in dense voxel embeddings.

%-------------------------------------------------------------------------
% \appendices
% Appendixes, if needed, appear before the acknowledgment.

%-------------------------------------------------------------------------
\section*{Acknowledgment}
We would like to give special thanks to Jonathan Zung, whose pioneering work on deep metric learning for connectomics back in 2016 had paved the road to the present work. We thank Kyle Willie, Ryan Willie, Ben Silverman and Selden Koolman for editing ground truth and proofreading segmentation. We are grateful to Nicholas Turner for sharing a custom ground truth editing tool, having valuable discussions, and proofreading the manuscript. We also thank Thomas Macrina and Nico Kemnitz for their helpful suggestions and proofreading efforts for the manuscript.

\bibliographystyle{IEEEtran}
\bibliography{IEEEabrv,main}

\begin{thebibliography}{10}
\providecommand{\url}[1]{#1}
\csname url@samestyle\endcsname
\providecommand{\newblock}{\relax}
\providecommand{\bibinfo}[2]{#2}
\providecommand{\BIBentrySTDinterwordspacing}{\spaceskip=0pt\relax}
\providecommand{\BIBentryALTinterwordstretchfactor}{4}
\providecommand{\BIBentryALTinterwordspacing}{\spaceskip=\fontdimen2\font plus
\BIBentryALTinterwordstretchfactor\fontdimen3\font minus
  \fontdimen4\font\relax}
\providecommand{\BIBforeignlanguage}[2]{{%
\expandafter\ifx\csname l@#1\endcsname\relax
\typeout{** WARNING: IEEEtran.bst: No hyphenation pattern has been}%
\typeout{** loaded for the language `#1'. Using the pattern for}%
\typeout{** the default language instead.}%
\else
\language=\csname l@#1\endcsname
\fi
#2}}
\providecommand{\BIBdecl}{\relax}
\BIBdecl

\bibitem{kornfeld2018progress}
J.~Kornfeld and W.~Denk, ``Progress and remaining challenges in high-throughput
  volume electron microscopy,'' \emph{Curr. Opin. Neurobiol.}, vol.~50, pp.
  261--267, 2018.

\bibitem{lee2019convolutional}
K.~Lee, N.~Turner, T.~Macrina, J.~Wu, R.~Lu, and H.~S. Seung, ``Convolutional
  nets for reconstructing neural circuits from brain images acquired by serial
  section electron microscopy,'' \emph{Curr. Opin. Neurobiol.}, vol.~55, pp.
  188--198, 2019.

\bibitem{zeng2017deepem3d}
T.~Zeng, B.~Wu, and S.~Ji, ``{DeepEM3D: approaching human-level performance on
  3D anisotropic EM image segmentation},'' \emph{Bioinformatics}, vol.~33,
  no.~16, pp. 2555--2562, 2017.

\bibitem{beier2017multicut}
T.~Beier \emph{et~al.}, ``Multicut brings automated neurite segmentation closer
  to human performance,'' \emph{Nat. Methods}, vol.~14, pp. 101--102, 2017.

\bibitem{lee2017superhuman}
K.~Lee, J.~Zung, P.~Li, V.~Jain, and H.~S. Seung, ``Superhuman accuracy on the
  {SNEMI3D} connectomics challenge,'' \emph{CoRR}, vol. abs/1706.00120, 2017.

\bibitem{funke2019large}
J.~{Funke} \emph{et~al.}, ``Large scale image segmentation with structured loss
  based deep learning for connectome reconstruction,'' \emph{IEEE Trans.
  Pattern Anal. Mach. Intell.}, vol.~41, no.~7, pp. 1669--1680, 2019.

\bibitem{bailoni2019generalized}
A.~Bailoni, C.~Pape, S.~Wolf, T.~Beier, A.~Kreshuk, and F.~A. Hamprecht, ``A
  generalized framework for agglomerative clustering of signed graphs applied
  to instance segmentation,'' \emph{CoRR}, vol. abs/1906.11713, 2019.

\bibitem{januszewski2016flood}
M.~Januszewski, J.~Maitin{-}Shepard, P.~Li, J.~Kornfeld, W.~Denk, and V.~Jain,
  ``Flood-filling networks,'' \emph{CoRR}, vol. abs/1611.00421, 2016.

\bibitem{januszewski2018high}
M.~Januszewski \emph{et~al.}, ``High-precision automated reconstruction of
  neurons with flood-filling networks,'' \emph{Nat. Methods}, vol.~15, no.~8,
  pp. 605--610, 2018.

\bibitem{meirovitch2016multipass}
Y.~Meirovitch \emph{et~al.}, ``A multi-pass approach to large-scale
  connectomics,'' \emph{CoRR}, vol. abs/1612.02120, 2016.

\bibitem{meirovitch2019cross}
Y.~Meirovitch, L.~Mi, H.~Saribekyan, A.~Matveev, D.~Rolnick, and N.~Shavit,
  ``Cross-classification clustering: An efficient multi-object tracking
  technique for 3-d instance segmentation in connectomics,'' in \emph{Proc.
  IEEE Conf. Comput. Vis. Pattern Recognit. (CVPR)}, 2019, pp. 8425--8435.

\bibitem{li2019automated}
P.~H. Li \emph{et~al.}, ``Automated reconstruction of a serial-section em
  drosophila brain with flood-filling networks and local realignment,''
  \emph{bioRxiv}, 2019.

\bibitem{scheffer2020connectome}
L.~K. Scheffer \emph{et~al.}, ``A connectome and analysis of the adult
  \textit{Drosophila} central brain,'' \emph{eLife}, vol.~9, p. e57443, 2020.

\bibitem{fathi2017semantic}
A.~Fathi \emph{et~al.}, ``Semantic instance segmentation via deep metric
  learning,'' \emph{CoRR}, vol. abs/1703.10277, 2017.

\bibitem{brabandere2017semantic}
B.~D. Brabandere, D.~Neven, and L.~V. Gool, ``Semantic instance segmentation
  with a discriminative loss function,'' \emph{CoRR}, vol. abs/1708.02551,
  2017.

\bibitem{luther2019learning}
K.~Luther and H.~S. Seung, ``Learning metric graphs for neuron segmentation in
  electron microscopy images,'' in \emph{Proc. IEEE Int. Symp. Biomed. Imaging
  (ISBI)}, 2019, pp. 244--248.

\bibitem{wolf2018mutex}
S.~Wolf \emph{et~al.}, ``The mutex watershed: Efficient, parameter-free image
  partitioning,'' in \emph{Comput. Vis. ECCV 2018}, 2018, pp. 571--587.

\bibitem{nunez-iglesias2013machine}
J.~Nunez-Iglesias, R.~Kennedy, T.~Parag, J.~Shi, and D.~B. Chklovskii,
  ``Machine learning of hierarchical clustering to segment 2d and 3d images,''
  \emph{PLoS One}, vol.~8, no.~8, pp. 1--11, 2013.

\bibitem{turaga2010convolutional}
S.~C. Turaga \emph{et~al.}, ``Convolutional networks can learn to generate
  affinity graphs for image segmentation,'' \emph{Neural Comput.}, vol.~22,
  no.~2, pp. 511--538, 2010.

\bibitem{zlateski2015image}
A.~Zlateski and H.~S. Seung, ``Image segmentation by size-dependent single
  linkage clustering of a watershed basin graph,'' \emph{CoRR}, vol.
  abs/1505.00249, 2015.

\bibitem{wolf2019semantic}
S.~Wolf \emph{et~al.}, ``The semantic mutex watershed for efficient bottom-up
  semantic instance segmentation,'' in \emph{Comput. Vis. ECCV 2020}, 2020, pp.
  208--224.

\bibitem{matejek2019biologically}
B.~Matejek, D.~Haehn, H.~Zhu, D.~Wei, T.~Parag, and H.~Pfister,
  ``Biologically-constrained graphs for global connectomics reconstruction,''
  in \emph{Proc. IEEE Conf. Comput. Vis. Pattern Recognit. (CVPR)}, 2019, pp.
  2089--2098.

\bibitem{krasowski2018neuron}
N.~E. {Krasowski}, T.~{Beier}, G.~W. {Knott}, U.~{Köthe}, F.~A. {Hamprecht},
  and A.~{Kreshuk}, ``Neuron segmentation with high-level biological priors,''
  \emph{IEEE Trans. Med. Imaging}, vol.~37, no.~4, pp. 829--839, 2018.

\bibitem{pape2019leveraging}
C.~Pape \emph{et~al.}, ``Leveraging domain knowledge to improve microscopy
  image segmentation with lifted multicuts,'' \emph{Front. Comput. Sci.},
  vol.~1, p.~6, 2019.

\bibitem{zung2017error}
J.~Zung, I.~Tartavull, K.~Lee, and H.~S. Seung, ``An error detection and
  correction framework for connectomics,'' in \emph{Adv. Neural Inf. Process.
  Syst. (NIPS)}, 2017, pp. 6818--6829.

\bibitem{dmitriev2018efficient}
K.~Dmitriev, T.~Parag, B.~Matejek, A.~Kaufman, and H.~Pfister, ``Efficient
  correction for em connectomics with skeletal representation,'' in
  \emph{BMVC}, 2018, p.~47.

\bibitem{schubert2019learning}
P.~J. Schubert, S.~Dorkenwald, M.~Januszewski, V.~Jain, and J.~Kornfeld,
  ``Learning cellular morphology with neural networks,'' \emph{Nat. Commun.},
  vol.~10, no.~1, p. 2736, 2019.

\bibitem{li2020neuronal}
H.~Li, M.~Januszewski, V.~Jain, and P.~H. Li, ``Neuronal subcompartment
  classification and merge error correction,'' in \emph{Med. Image Comput.
  Comput. Assist. Interv. (MICCAI)}, 2020, pp. 88--98.

\bibitem{lee2019learning}
K.~Lee, R.~Lu, K.~Luther, and H.~S. Seung, ``Learning dense voxel embeddings
  for 3d neuron reconstruction,'' \emph{CoRR}, vol. abs/1909.09872, 2019.

\bibitem{han2020occuseg}
L.~Han, T.~Zheng, L.~Xu, and L.~Fang, ``Occuseg: Occupancy-aware 3d instance
  segmentation,'' in \emph{Proc. IEEE Conf. Comput. Vis. Pattern Recognit.
  (CVPR)}, 2020, pp. 2940--2949.

\bibitem{harley2015learning}
A.~W. Harley, K.~G. Derpanis, and I.~Kokkinos, ``Learning dense convolutional
  embeddings for semantic segmentation,'' \emph{CoRR}, vol. abs/1511.04377,
  2015.

\bibitem{newell2017associative}
A.~Newell, Z.~Huang, and J.~Deng, ``Associative embedding: End-to-end learning
  for joint detection and grouping,'' in \emph{Adv. Neural Inf. Process. Syst.
  (NIPS)}, 2017, pp. 2277--2287.

\bibitem{pham2019jsis3d}
Q.-H. Pham, T.~Nguyen, B.-S. Hua, G.~Roig, and S.-K. Yeung, ``Jsis3d: Joint
  semantic-instance segmentation of 3d point clouds with multi-task pointwise
  networks and multi-value conditional random fields,'' in \emph{Proc. IEEE
  Conf. Comput. Vis. Pattern Recognit. (CVPR)}, 2019, pp. 8827--8836.

\bibitem{lahoud2019instance}
J.~Lahoud, B.~Ghanem, M.~Pollefeys, and M.~R. Oswald, ``3d instance
  segmentation via multi-task metric learning,'' in \emph{Proc. IEEE Int. Conf.
  Comput. Vis. (ICCV)}, 2019, pp. 9256--9266.

\bibitem{halupka2019deep}
K.~{Halupka}, R.~{Garnavi}, and S.~{Moore}, ``Deep semantic instance
  segmentation of tree-like structures using synthetic data,'' in \emph{2019
  IEEE Winter Conference on Applications of Computer Vision (WACV)}, 2019, pp.
  1713--1722.

\bibitem{tian2019learning}
Z.~Tian \emph{et~al.}, ``Learning shape-aware embedding for scene text
  detection,'' in \emph{Proc. IEEE Conf. Comput. Vis. Pattern Recognit.
  (CVPR)}, 2019, pp. 4229--4238.

\bibitem{kong2018recurrent}
S.~Kong and C.~C. Fowlkes, ``Recurrent pixel embedding for instance grouping,''
  in \emph{Proc. IEEE Conf. Comput. Vis. Pattern Recognit. (CVPR)}, 2018, pp.
  2858--2866.

\bibitem{xie2019object}
C.~Xie, Y.~Xiang, Z.~Harchaoui, and D.~Fox, ``Object discovery in videos as
  foreground motion clustering,'' in \emph{Proc. IEEE Conf. Comput. Vis.
  Pattern Recognit. (CVPR)}, 2019, pp. 9986--9995.

\bibitem{payer2018instance}
C.~Payer, D.~{\v{S}}tern, T.~Neff, H.~Bischof, and M.~Urschler, ``Instance
  segmentation and tracking with cosine embeddings and recurrent hourglass
  networks,'' in \emph{Med. Image Comput. Comput. Assist. Interv. (MICCAI)},
  2018, pp. 3--11.

\bibitem{payer2019segmenting}
C.~Payer, D.~\v{S}tern, M.~Feiner, H.~Bischof, and M.~Urschler, ``Segmenting
  and tracking cell instances with cosine embeddings and recurrent hourglass
  networks,'' \emph{Med. Image Anal.}, vol.~57, pp. 106 -- 119, 2019.

\bibitem{chen2019instance}
L.~Chen, M.~Strauch, and D.~Merhof, ``Instance segmentation of biomedical
  images with an object-aware embedding learned with local constraints,'' in
  \emph{Med. Image Comput. Comput. Assist. Interv. (MICCAI)}, 2019, pp.
  451--459.

\bibitem{Konopczynski2018instance}
T.~K. Konopczynski, T.~Kr{\"{o}}ger, L.~Zheng, and J.~Hesser, ``Instance
  segmentation of fibers from low resolution {CT} scans via 3d deep embedding
  learning,'' in \emph{BMVC}, 2018, p. 268.

\bibitem{ronneberger2015unet}
O.~Ronneberger, P.Fischer, and T.~Brox, ``U-net: Convolutional networks for
  biomedical image segmentation,'' in \emph{Med. Image Comput. Comput. Assist.
  Interv. (MICCAI)}, 2015, pp. 234--241.

\bibitem{lichtman2014}
J.~W. Lichtman, H.~Pfister, and N.~Shavit, ``The big data challenges of
  connectomics,'' \emph{Nat. Neurosci.}, vol.~17, no.~11, pp. 1448--54, 2014.

\bibitem{kasthuri2015saturated}
N.~Kasthuri \emph{et~al.}, ``Saturated reconstruction of a volume of
  neocortex,'' \emph{Cell}, vol. 162, no.~3, pp. 648--661, 2015.

\bibitem{odena2016deconvolution}
\BIBentryALTinterwordspacing
A.~Odena, V.~Dumoulin, and C.~Olah, ``Deconvolution and checkerboard
  artifacts,'' \emph{Distill}, 2016. [Online]. Available:
  \url{http://distill.pub/2016/deconv-checkerboard}
\BIBentrySTDinterwordspacing

\bibitem{reddi2018}
S.~J. Reddi, S.~Kale, and S.~Kumar, ``On the convergence of adam and beyond,''
  in \emph{ICLR}, 2018.

\bibitem{kingma2015}
D.~P. Kingma and J.~Ba, ``Adam: A method for stochastic optimization,'' in
  \emph{ICLR}, 2015.

\bibitem{arganda-carreras2015crowdsourcing}
I.~Arganda-Carreras \emph{et~al.}, ``Crowdsourcing the creation of image
  segmentation algorithms for connectomics,'' \emph{Front. Neuroanat.}, vol.~9,
  p. 142, 2015.

\bibitem{helmstaedter2013cellular}
M.~Helmstaedter, ``Cellular-resolution connectomics: challenges of dense neural
  circuit reconstruction,'' \emph{Nat. Methods}, vol.~10, no.~6, pp. 501--507,
  2013.

\bibitem{matveev2017multicore}
A.~Matveev \emph{et~al.}, ``A multicore path to connectomics-on-demand,''
  \emph{SIGPLAN Not.}, vol.~52, no.~8, pp. 267--281, 2017.

\bibitem{gonda2021consistent}
F.~Gonda, D.~Wei, and H.~Pfister, ``Consistent recurrent neural networks for 3d
  neuron segmentation,'' in \emph{Proc. IEEE Int. Symp. Biomed. Imaging
  (ISBI)}, 2021.

\bibitem{wu2018group}
Y.~Wu and K.~He, ``Group normalization,'' in \emph{Comput. Vis. ECCV 2018},
  2018, pp. 3--19.

\bibitem{ulyanov2016instance}
D.~Ulyanov, A.~Vedaldi, and V.~S. Lempitsky, ``Instance normalization: The
  missing ingredient for fast stylization,'' \emph{CoRR}, vol. abs/1607.08022,
  2016.

\end{thebibliography}

% \end{document}

\clearpage

\def\BibTeX{{\rm B\kern-.05em{\sc i\kern-.025em b}\kern-.08em
    T\kern-.1667em\lower.7ex\hbox{E}\kern-.125emX}}
% \markboth{\journalname, VOL. XX, NO. XX, XXXX 2021}
% {Lee \MakeLowercase{\textit{et al.}}: Preparation of Papers for IEEE TRANSACTIONS and JOURNALS (July 2021)}

% Subfigure
% \usepackage[caption=false,font=footnotesize]{subfig}

% \begin{document}

% \title{Supplementary Notes for ``Learning and Segmenting Dense Voxel Embeddings for 3D Neuron Reconstruction''}

% \author{Kisuk Lee, Ran Lu, Kyle Luther, and H. Sebastian Seung}

% \maketitle

\newcommand{\beginsupplement}{%
    \setcounter{table}{0}
    \renewcommand{\thetable}{S\arabic{table}}%
    \setcounter{figure}{0}
    \renewcommand{\thefigure}{S\arabic{figure}}%
    \setcounter{section}{0}
    \renewcommand{\thesection}{S\arabic{section}}%
}

\beginsupplement

\section{SNEMI3D Experiments}

We trained the embedding net strictly under the SNEMI3D challenge setup, and segmented the test volume with our proposed method. We submitted our segmentation to the challenge leaderboard, allowing for objective comparison with other methods. 

In all submissions, we used the same set of postprocessing parameters as in the main text, except for $\theta_\text{mask} = 0.5$. For both training and inference, we used $2\times$ in-plane downsampled images. We used the top 75 sections of the SNEMI3D training volume for training, and the bottom 25 sections for validation. Since the embedding net produces an output patch smaller than its input patch, extra context is required at the edge of the volume. We used mirror-padding to extend the test volume to provide extra context for inference. Prior to submission, the segmentation was dilated in 2D until no background voxels remain, and then upsampled with nearest neighbor upsampling to recover its original size. 

We first started with the same network architecture as in the main text. The result was evaluated with an adapted Rand error of 0.0319, which outperformed the two recently published methods \cite{meirovitch2019cross,gonda2021consistent}, but fell behind the two topmost entries \cite{januszewski2018high, lee2017superhuman} (see Table~\ref{tab:snemi3d}). 

Qualitative inspection of our initial segmentation revealed two major problems. First, we found several merge errors near the edge of the volume, mainly caused by the systematic degradation of the embeddings and affinities near the volume edge. This is presumably due to the use of mirror-padding, which creates a biologically implausible scene that is never seen during training. Second, we found a different set of merge errors that seem to occur by generalization failure. This may have happened because the amount of training data (78.6 million voxels at the original resolution) is far less than that used in the main text (226.5 million voxels at the original resolution).

To address these issues, we explored two variations in the network architecture: the use of (1) {\it Group Normalization} (GN, \cite{wu2018group}) instead of {\it Instance Normalization} (IN,\cite{ulyanov2016instance}), and (2) smaller training/inference patches. GN is known to act as a strong regularizer and help achieving better generalization performance than IN \cite{wu2018group}. We also hypothesized that using smaller {\it training} patches may benefit generalization, because it increases the effective size of the training set, decreases correlation between patches, and reduces the complexity of the input scene. Additionally, using smaller {\it inference} patches may make inference more robust to the biologically implausible scene created by mirror-padding, because smaller context may make local statistics more similar to that of the real image.

As can be seen in Table~\ref{tab:snemi3d}, using GN and smaller patch size ($64 \times 64 \times 20$) reduced the error from 0.0319 to 0.0265 and 0.0262 respectively, both outperforming the flood-filling net (FFN, \cite{januszewski2018high}) and the mean affinity agglomeration (MAA) result of \cite{lee2017superhuman}. Combining both further reduced the error down to 0.0256, which is only marginally higher than the leading entry's error of 0.0249 \cite{lee2017superhuman}. 

Qualitative comparison of our best submission and the SNEMI3D's leading entry \cite{lee2017superhuman} revealed that ours has a substantially less number of split errors on thin neurites (it reduced 15 dendritic spine splits and 14 axon splits), but also has a couple of more merge errors. This suggests that such split errors may have been under reflected in the Rand error because of its overdependence on the size of erroneous objects. In other words, a small number of bigger merge errors could easily overwhelm the effect of a large number of smaller split errors. 

\begin{table*}[ht!]
\begin{center}
\caption{SNEMI3D Results}\label{tab:snemi3d}
\begin{tabular}{|l|l|c|c|c|}
\hline
Group Name & Method & Group Norm \cite{wu2018group} & Patch Size & Rand Error$\downarrow$\\
\hline
\hline
PNI & Watershed~\cite{zlateski2015image} + TTA ($16\times$) \cite{zeng2017deepem3d,lee2017superhuman} & & & 0.0249 \\
\hline
{\bf Ours} & Metric Learning + MWS \cite{wolf2018mutex} + MEA & \checkmark & $64\times64\times20$ & {\bf 0.0256} \\
\hline
{\bf Ours} & Metric Learning + MWS \cite{wolf2018mutex} + MEA & & $64\times64\times20$ & {\bf 0.0262} \\
\hline
{\bf Ours} & Metric Learning + MWS \cite{wolf2018mutex} + MEA & \checkmark & $128\times128\times20$ & {\bf 0.0265} \\
\hline
GAIP & Flood-Filling Networks \cite{januszewski2016flood,januszewski2018high} & & & 0.0291 \\
\hline
PNI & Watershed~\cite{zlateski2015image} + MAA \cite{lee2017superhuman} & & & 0.0314 \\
\hline
{\bf Ours} & Metric Learning + MWS \cite{wolf2018mutex} + MEA & & $128\times128\times20$ & {\bf 0.0319} \\
\hline
% VIDAR & & & & 0.0342 \\
% \hline
S\&T & STRU-Net \cite{gonda2021consistent} & & & 0.0351 \\
\hline
{\bf Ours} & Metric Learning + MWS \cite{wolf2018mutex} & & $64\times64\times20$ & {\bf 0.0379} \\
\hline
{\bf Ours} & Metric Learning + MWS \cite{wolf2018mutex} & \checkmark & $64\times64\times20$ & {\bf 0.0386} \\
\hline
CCG & Cross-Classification Clustering \cite{meirovitch2019cross} & & & 0.0410 \\
\hline
{\bf Ours} & Metric Learning + MWS \cite{wolf2018mutex} & \checkmark & $128\times128\times20$ & {\bf 0.0454} \\
\hline
{\bf Ours} & Metric Learning + MWS \cite{wolf2018mutex} & & $128\times128\times20$ & {\bf 0.0456} \\
\hline
% LZL-USTC & & & & 0.0461 \\
% \hline
% CS17 & & & & 0.0469 \\
% \hline
% VCG & & & & 0.0584 \\
% \hline
\hline
Human & & & & 0.0600 \\
\hline
\hline
DIVE & DeepEM3D \cite{zeng2017deepem3d} & & & 0.0602 \\
\hline
IAL & Lifted Multicut \cite{beier2017multicut} & & & 0.0656 \\
\hline
\multicolumn{5}{p{300pt}}{$\downarrow$ The lower, the better.}\\
\multicolumn{5}{p{360pt}}{Abbreviations: Mean Affinity Agglomeration (MAA), Test-Time Augmentation (TTA), Mutex Watershed (MWS), Mean Embedding agglomeration (MEA)}\\
\end{tabular}
\end{center}
\end{table*}

\section{Notes on postprocessing parameters}

\subsection{$\theta_\text{mask}$}
The $\theta_\text{mask}$ parameter is a threshold for the real-valued background mask predicted by the embedding net to produce a binary background mask, which is used for excluding background voxels (nodes) and their incident affinities (edges) from the metric graph. $\theta_\text{mask}$ effectively controls the level of how conservatively we restrict the metric graph to foreground voxels by removing the potentially noisy embeddings and affinities near the background. When too liberal, mergers can occur through the uncertainty near the background. When too conservative, connectivity of the thin neurites may be altered.

To measure sensitivity of our system to varying $\theta_\text{mask}$, we performed 3-fold cross-validation. To construct three folds, we split AC3/AC4 into three same-sized volumes: AC4, the top 100 slices of AC3, and the bottom 100 slices of AC3. Besides the three folds, we used the middle 56 slices of AC3 as an extra training set in all cases.

Table~\ref{tab:mask} summarizes the 3-fold cross-validation result for $\theta_\text{mask}$. For each value of $\theta_\text{mask}$ on each fold, we optimized $\theta_d$ to obtain the best VI. Overall, $\theta_\text{mask}$ around 0.5 was found to be optimal, and segmentation accuracy measured by VI tends to decrease as $\theta_\text{mask}$ gets farther away from this optimal value.

\begin{table}[]
    \centering
    \caption{3-fold cross-validation for $\theta_\text{mask}$}
    \begin{tabular}{|c|c|c|c|c|}
        \hline
        $\theta_\text{mask}$ & Fold 1 & Fold 2 & Fold3 & Mean $\pm$ S.E. \\
        \hline
        \hline
        0.1 & 0.1243 & 0.1338 & 0.1159 & 0.1245 $\pm$ 0.0052 \\
        \hline
        0.2 & 0.0746 & 0.1094 & 0.0818 & 0.0886 $\pm$ 0.0106 \\
        \hline
        0.3 & 0.0601 & 0.1625 & 0.0984 & 0.1070 $\pm$ 0.0299 \\
        \hline
        0.4 & 0.0566 & {\bf 0.0628} & 0.0775 & 0.0656 $\pm$ 0.0062 \\
        \hline
        0.5 & 0.0416 & 0.0681 & {\bf 0.0763} & {\bf 0.0620} $\pm$ 0.0105 \\
        \hline
        0.6 & {\bf 0.0409} & 0.0689 & 0.0781 & 0.0626 $\pm$ 0.0112 \\
        \hline
        0.7 & 0.0469 & 0.1345 & 0.0877 & 0.0897 $\pm$ 0.0253 \\
        \hline
        0.8 & 0.0465 & 0.0651 & 0.0882 & 0.0666 $\pm$ 0.0121 \\
        \hline
        0.9 & 0.0680 & 0.0847 & 0.0907 & 0.0811 $\pm$ 0.0068 \\
        \hline
    \end{tabular}
    \label{tab:mask}
\end{table}

\subsection{$\theta_d$}
We performed a similar sensitivity analysis for $\theta_d$ with 3-fold cross-validation. Table~\ref{tab:delta} summarizes the 3-fold cross-validation result for $\theta_d$, with $\theta_\text{mask}$ fixed to 0.5. Overall, the optimal range for $\theta_d$ was $\theta_d \in [1.0, 1.5)$. We found that the optimal value for $\theta_d$ on fold 2 is unusual and deviates from the overall optimal range. This was caused by a complication of merge errors. Although a {\it local} decision to agglomerate two segments is correct, {\it global} segmentation accuracy could get worsen if either of the segments involves a merge error. In other words, agglomerating two segments may correct a split error while amplifying an existing merge error. This property is generally relevant to any kind of error correction, and should be considered carefully when designing a downstream system for human proofreading.

\begin{table}[]
    \centering
    \caption{3-fold cross-validation for $\theta_d$ with $\theta_\text{mask} = 0.5$}
    \begin{tabular}{|c|c|c|c|c|}
        \hline
        $\theta_\text{mask}$ & Fold 1 & Fold 2 & Fold3 & Mean $\pm$ S.E. \\
        \hline
        \hline
        0.50 & 0.0586 & {\bf 0.0681} & 0.0885 & 0.0718 $\pm$ 0.0088 \\
        \hline
        0.75 & 0.0558 & {\bf 0.0681} & {\bf 0.0763} & 0.0668 $\pm$ 0.0060 \\
        \hline
        1.00 & {\bf 0.0416} & 0.0729 & {\bf 0.0763} & {\bf 0.0636} $\pm$ 0.0110 \\
        \hline
        1.25 & {\bf 0.0416} & 0.0729 & {\bf 0.0763} & {\bf 0.0636} $\pm$ 0.0110 \\
        \hline
        1.50 & {\bf 0.0416} & 0.0728 & 0.0778 & 0.0641 $\pm$ 0.0113 \\
        \hline
        1.75 & 0.0465 & 0.0728 & 0.0778 & 0.0657 $\pm$ 0.0097 \\
        \hline
        2.00 & 0.0465 & 0.0728 & 0.0778 & 0.0657 $\pm$ 0.0097 \\
        \hline
        2.25 & 0.0502 & 0.0728 & 0.0778 & 0.0670 $\pm$ 0.0085 \\
        \hline
        2.50 & 0.0502 & 0.0728 & 0.0778 & 0.0670 $\pm$ 0.0085 \\
        \hline
    \end{tabular}
    \label{tab:delta}
\end{table}

\section{Inference benchmark}

Table~\ref{tab:time} shows the decomposition of runtime for processing a $512 \times 512 \times 100$ volume at the voxel resolution of $12 \times 12 \times 29$ nm$^3$, averaged over six volumes. The benchmark was performed on a workstation equipped with Intel\textregistered~Core\texttrademark~i7-9700K CPU @ 3.60GHz and Nvidia GeForce RTX 2080 Ti. As can be seen in Table~\ref{tab:time}, the embedding net inference and the Mutex Watershed collectively dominate the total runtime (97.61\% on average).

% \begin{table*}[]
%     \centering
%     \caption{Inference time decomposition}
%     \begin{tabular}{|r|c|c|c|c|c|c|c|c|}
%         \hline
%          & AC4 & AC3$^1$ & AC3$^2$ & E1 & E2 & E3 & Mean $\pm$ S.E. & Fraction  \\
%          \hline
%          \hline
%          Embedding net inference (s) & 191.52 & 193.54 & 185.47 & 187.49 & 189.50 & 191.52 & 189.84 $\pm$ 1.21 & 53.09\% \\
%          \hline
%          Mutex Watershed (s) & 146.20 & 194.27 & 153.80 & 156.06 & 135.40 & 169.42 & 159.19 $\pm$ 8.38 & 44.52\% \\
%          \hline
%          Self-contact split detection (s) & 6.44 & 5.94 & 6.04 & 7.01 & 7.32 & 6.08 & 6.47 $\pm$ 0.23 & 1.81\% \\
%          \hline
%          Mean embedding agglomeration (s) & 0.85 & 1.51 & 2.05 & 2.57 & 2.67 & 2.65 & 2.05 $\pm$ 0.30 & 0.57\% \\
%          \hline
%          \hline
%          Total (s) & 345.01 & 395.25 & 347.36 & 353.13 & 334.89 & 369.67 & 357.55 $\pm$ 8.87 & 100.00\% \\
%          \hline
%     \end{tabular}
%     \label{tab:time}
% \end{table*}

\begin{table}[]
    \centering
    \caption{Inference time decomposition}
    \begin{tabular}{|r|c|c|}
        \hline
         & Runtime (s) & Fraction (\%)  \\
         \hline
         \hline
         Embedding net inference & 189.84 & 53.09 \\
         \hline
         Mutex Watershed & 159.19 & 44.52 \\
         \hline
         Self-contact split detection & 6.47 & 1.81 \\
         \hline
         Mean embedding agglomeration & 2.05 & 0.57 \\
         \hline
         \hline
         Total & 357.55 & 100.00 \\
         \hline
    \end{tabular}
    \label{tab:time}
\end{table}

% \bibliographystyle{IEEEtran}
% \bibliography{IEEEabrv,supp}

% \end{document}

\clearpage

% \documentclass[journal,web]{ieeecolor}
% \documentclass{article}
% \usepackage{tmi}
% \usepackage{cite}
% \usepackage{amsmath,amssymb,amsfonts}
% \usepackage{algorithmic}
% \usepackage{graphicx}
% \usepackage{textcomp}
% \usepackage{hyperref}

% Subfigure
% \usepackage[caption=false,font=footnotesize]{subfig}

% \begin{document}

\newcommand{\beginsupplement}{%
    \setcounter{table}{0}
    \renewcommand{\thetable}{S\arabic{table}}%
    \setcounter{figure}{0}
    \renewcommand{\thefigure}{S\arabic{figure}}%
}

\beginsupplement

\begin{figure*}[!ht]
\begin{center}
\includegraphics[width=0.85\textwidth]{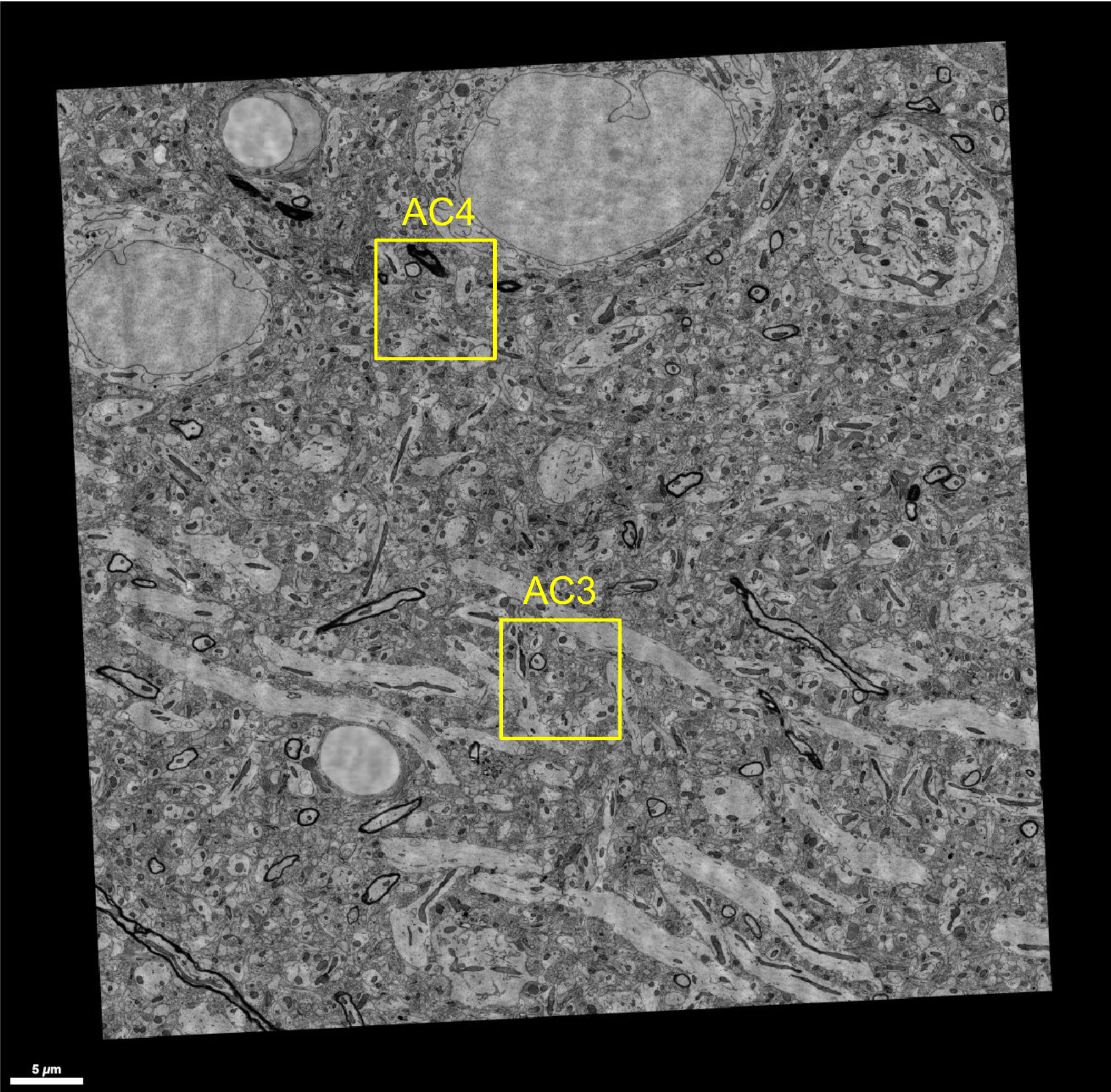}
\caption{Location of AC3/AC4 in the full dataset of Kasthuri \textit{et al.} (2015). Shown here is the section at $z = 1099$. The full dataset is accessible via \url{https://neurodata.io/data/kasthuri15/}. AC3's bounding box in the full dataset is (10944,17424,999)--(12992,19472,1255), and AC4's bounding box is (8800,10880,1099)--(10848,12928,1199). Note that the original images were acquired at $3\times3\times29$ nm$^3$ voxel resolution. As a result, the bounding boxes are $2\times$ larger in $x$ and $y$ dimension than the AC3/AC4 dataset, which was annotated at $6\times6\times29$ nm$^3$ voxel resolution.} \label{fig:AC3AC4}
\end{center}
\end{figure*}

\begin{figure*}[!ht]
\begin{center}
\includegraphics[width=0.85\linewidth]{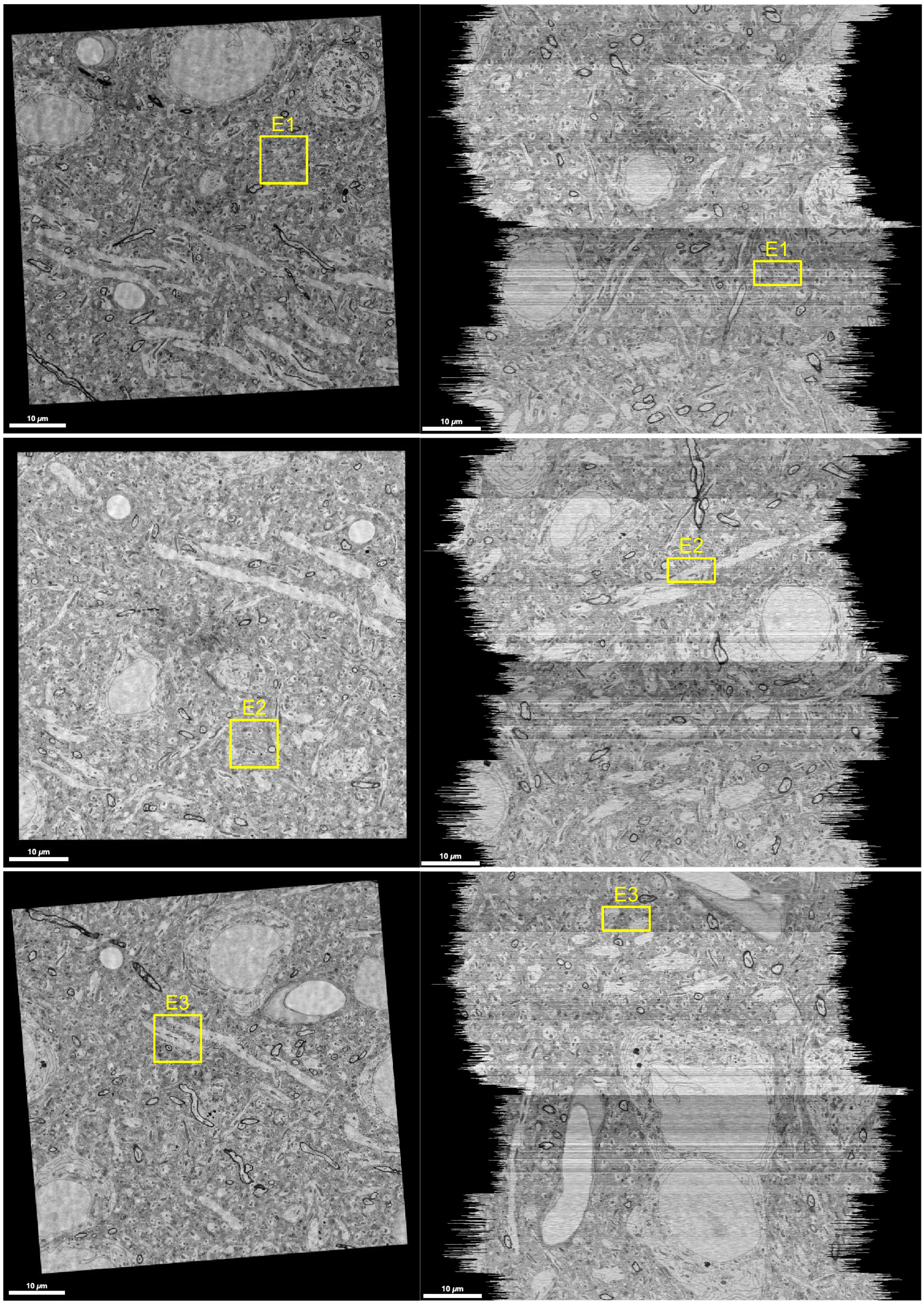}
\caption{Location of the extra test volumes E1--E3 in the full dataset of Kasthuri \textit{et al.} (2015). {\bf Left column:} $xy$ view. {\bf Right column:} $xz$ reslice view. E1's bounding box: (14300,12780,1099)--(16348,14828,1199). E2's bounding box: (10600,16200,516)--(12648,18248,616). E3's bounding box: (7800,9400,150)--(9848,11448,250).} \label{fig:E1-E3_location}
\end{center}
\end{figure*}

\begin{figure*}[!ht]
\begin{center}
\includegraphics[width=0.95\textwidth]{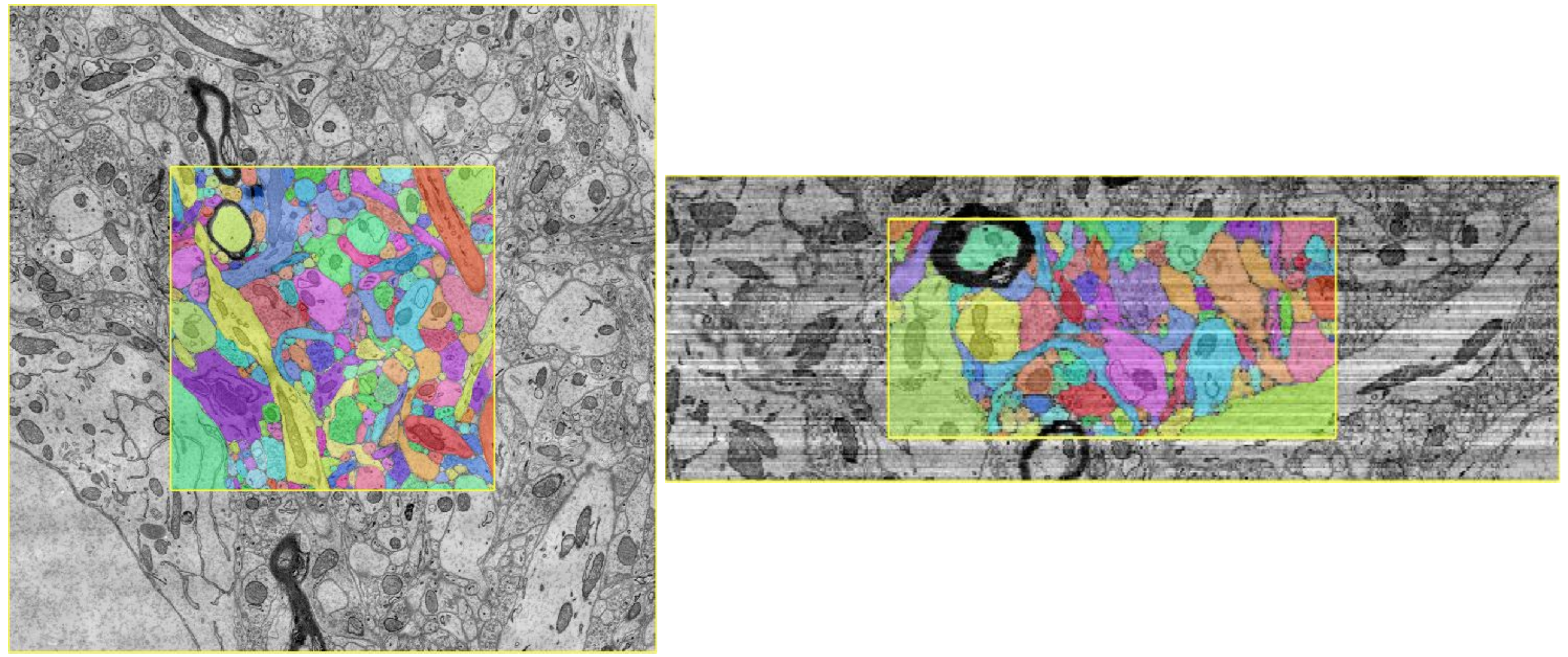}
\caption{Visualization of AC4 with surrounding image context. {\bf Left:} $xy$ view. {\bf Right:} $xz$ reslice view. We used the entire AC4 for training. In the original AC4 annotation, myelin sheath is not labeled separately. We additionally labeled myelin sheaths as separate objects.} \label{fig:AC4}
\end{center}
\end{figure*}

\begin{figure*}[!ht]
\begin{center}
\includegraphics[width=0.95\textwidth]{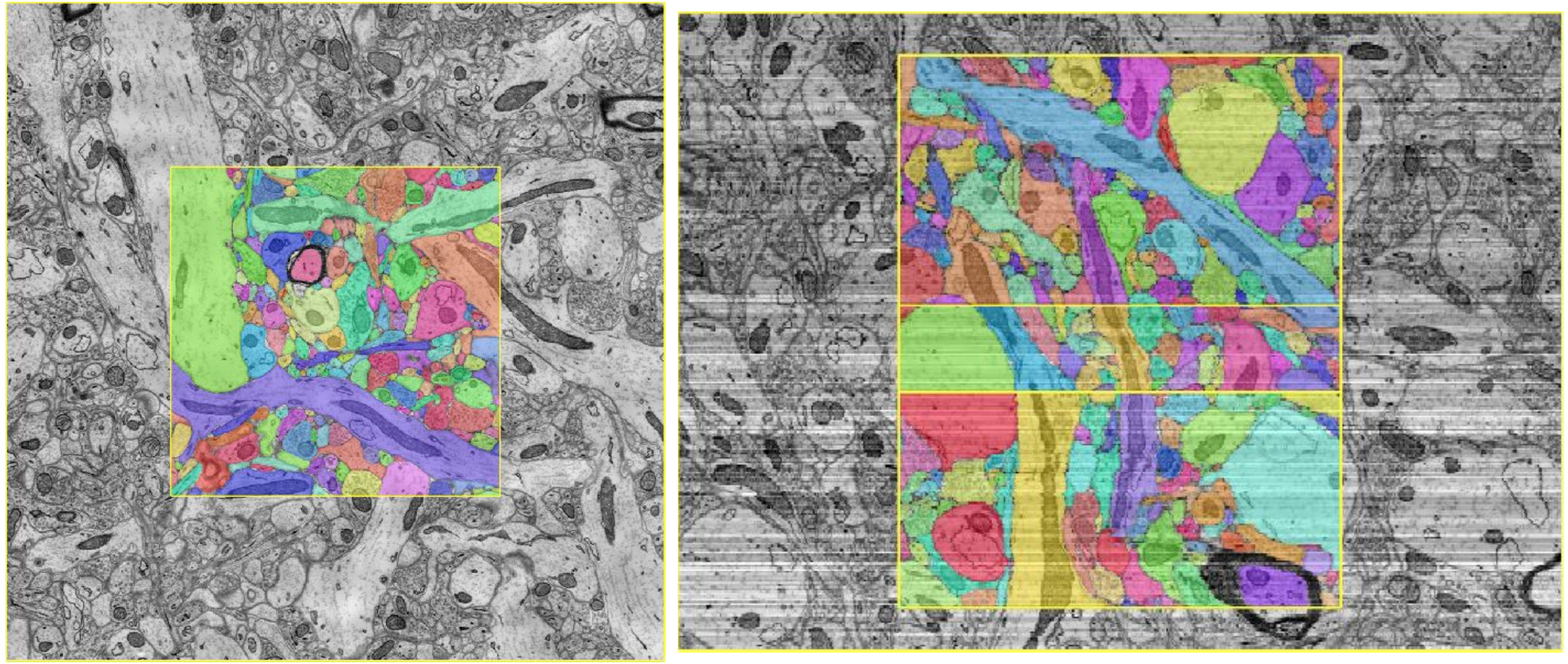}
\caption{Visualization of AC3 with surrounding image context. {\bf Left:} $xy$ view. {\bf Right:} $xz$ reslice view. We used the top 116 slices for training, the middle 40 slices for validation, and the bottom 100 slices for testing. Note that the volume shown here is a flipped version of the downloadable AC3 volume, so the top/middle/bottom relationship of the training/validation/test split here is different from the description in the main text. In the original AC3 annotation, myelin sheath is not labeled separately. We additionally labeled myelin sheaths as separate objects, and treated them differently in $\mathcal{L}_\text{embedding}$ and $\mathcal{L}_\text{background}$. Specifically, we treated myelin sheaths as background in $\mathcal{L}_\text{background}$, whereas they were treated as foreground objects in $\mathcal{L}_\text{embedding}$.} \label{fig:AC3_split}
\end{center}
\end{figure*}

\begin{figure*}[!ht]
\begin{center}
\includegraphics[width=0.5\textwidth]{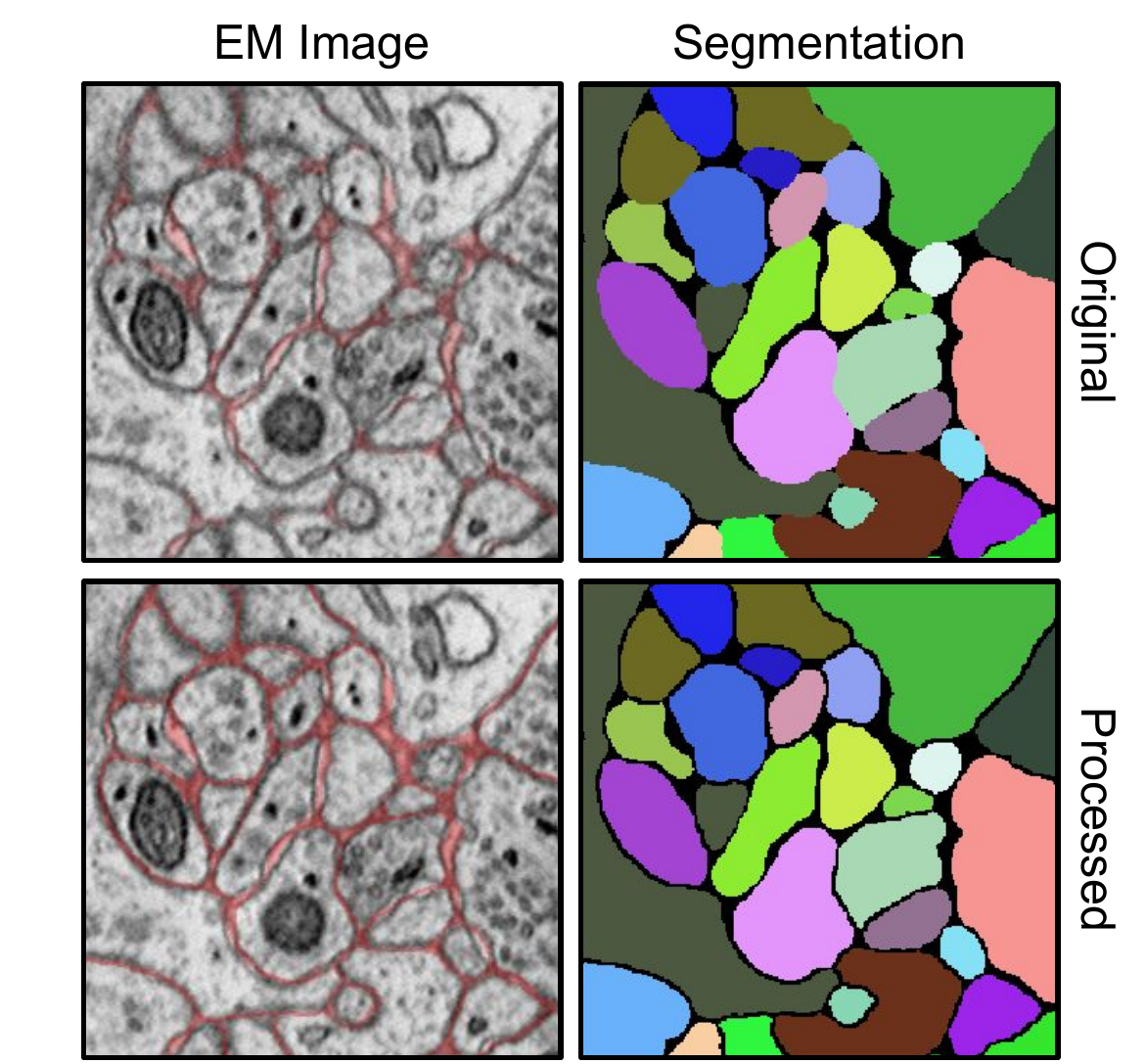}
\caption{Data preprocessing. For input image, we linearly rescaled the original integer-valued intensity in $[0,255]$ to the real-valued intensity in $[0,1]$. For background mask prediction, we applied a $3\times3\times1$ ``background-augmenting'' kernel to the ground truth segmentation such that any voxel whose $3\times3\times1$ neighbors (including the voxel itself) contain more than one \emph{positive} segment ID (\emph{zero} is reserved for background) was additionally marked as background.} \label{fig:preprocessing}
\end{center}
\end{figure*}

\begin{figure*}[!ht]
\begin{center}
\includegraphics[width=0.8\textwidth]{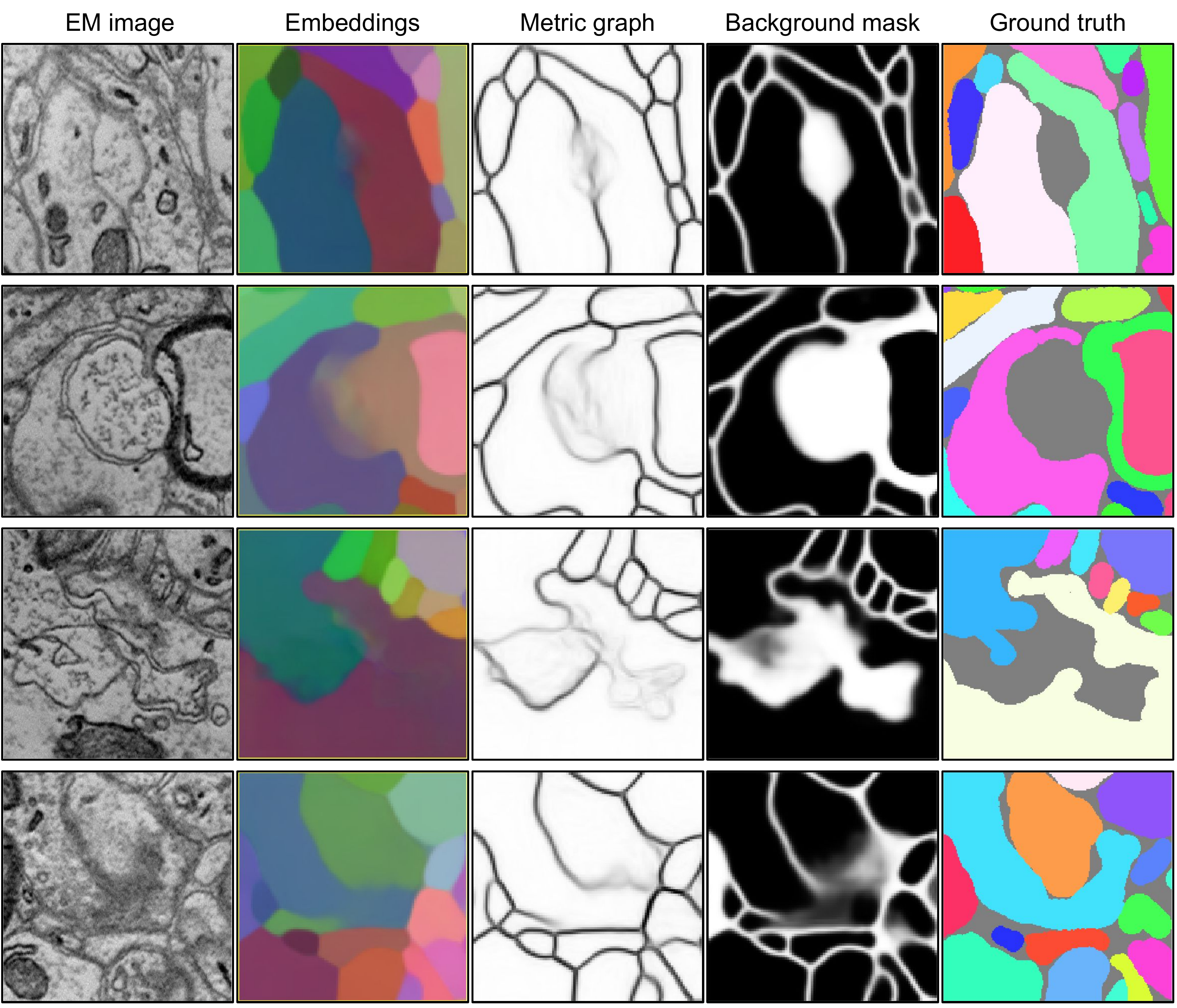}
\caption{Noisy embeddings resulted from the exclusion of background voxels from $\mathcal{L}_\text{embedding}$. {\bf Top row:} extracellular space between adjacent objects is enlarged, forming an object-like structure that is labeled as background. {\bf Second/third rows:} endoplasmic reticulum from each side of adjacent objects form a complex structure that is labeled as background. Note that we treated the myelin sheath as background in $\mathcal{L}_\text{background}$ (second row, fourth column), whereas it was treated as a foreground object in $\mathcal{L}_\text{embedding}$ (second row, second column). {\bf Bottom row:} diffuse boundaries at the synaptic interface that is parallel to the imaging plane. Here the predicted background mask is also noisy (bottom row, fourth column), thus requiring long-range affinities to be included as repulsive constraints during clustering. To visualize metric graph (third column), we used $\min(a_x, a_y)$ where $a_x$/$a_y$ are $x$/$y$ nearest neighbor metric-derived affinities.} \label{fig:background}
\end{center}
\end{figure*}

\begin{figure*}[!ht]
\begin{center}
\includegraphics[width=0.8\textwidth]{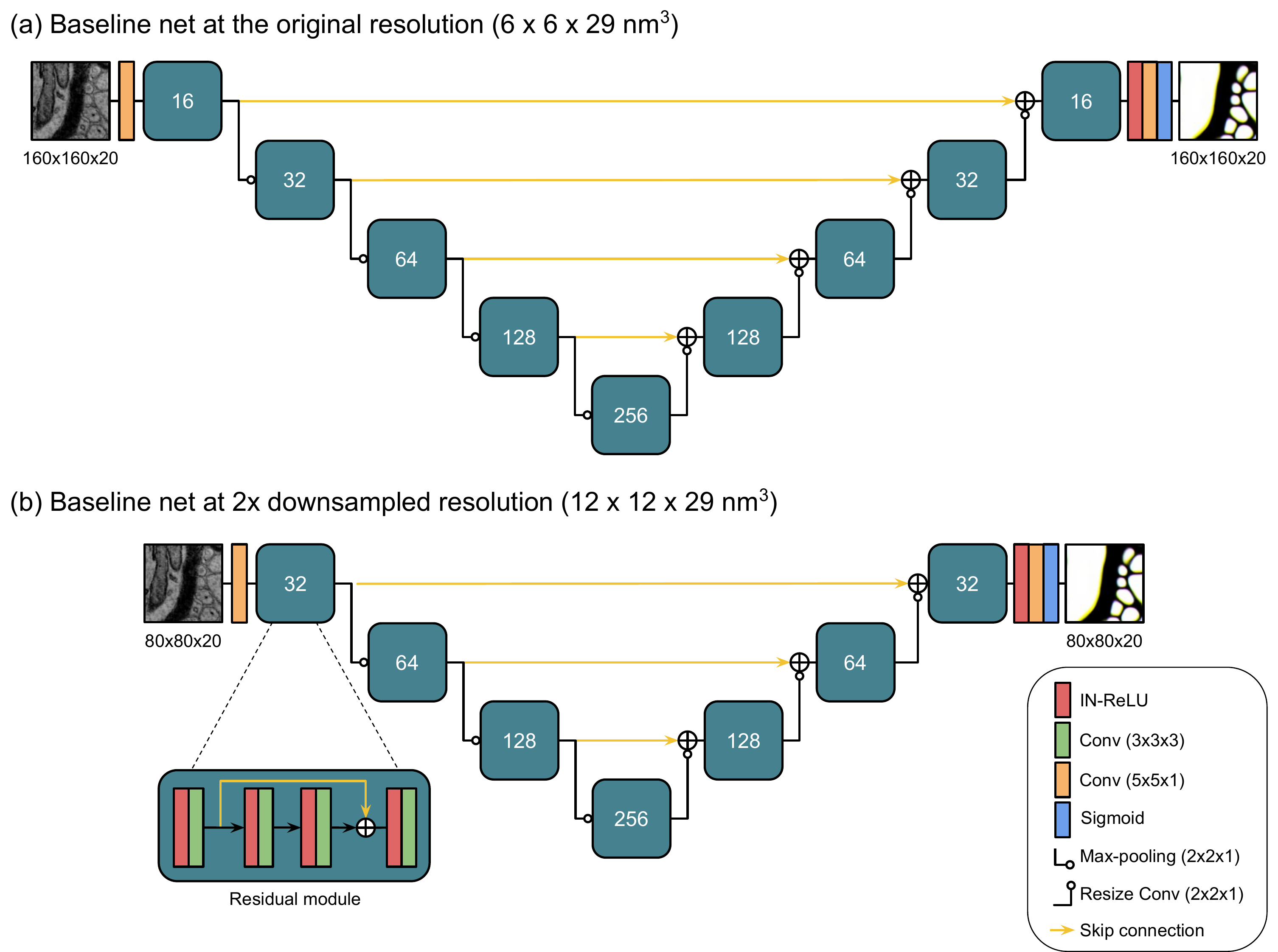}
\caption{Baseline net architecture. (a) We used a training patch of $160\times160\times20$ and an inference patch of $320\times320\times20$ voxels at the original resolution ($6\times6\times29$ nm$^3$). (b) For $2\times$ in-plane-downsampled images ($12\times12\times29$ nm$^3$), we accordingly halved the training and inference patch sizes in $x$ and $y$ dimension. Using larger inference patch was found to be effective at suppressing prediction noises in the large intracellular regions, although there was no significant difference in segmentation accuracy after postprocessing. Besides predicting nearest neighbor affinities as a primary target, we used long-range affinities as an auxiliary target during training. For generating long-range affinity maps, we used the following offset vectors: $(-4,0,0)$, $(-8,0,0)$, $(-24,0,0)$, $(0,-4,0)$, $(0,-8,0)$, $(0,-24,0)$, $(0,0,-2)$, $(0,0,-3)$, and $(0,0,-4)$ at the original resolution. We halved the $x$ and $y$ offsets for $2\times$ in-plane-downsampled images. As a result, the baseline nets produce 12 output channels, three for nearest neighbor and nine for long-range affinities. Here the dimensionality of output channels is omitted for brevity and only spatial dimensions are displayed. The number inside the residual module represents the width (number of feature maps) of the module. For upsampling, we used the bilinear \emph{resize convolution}, i.e., bilinear upsampling followed by a pointwise ($1\times1\times1$) convolution. Abbreviations: Instance Normalization (IN), rectified linear unit (ReLU), convolution (Conv).} \label{fig:baseline_net}
\end{center}
\end{figure*}

\begin{figure*}[t]
\begin{center}
\includegraphics[width=0.8\textwidth]{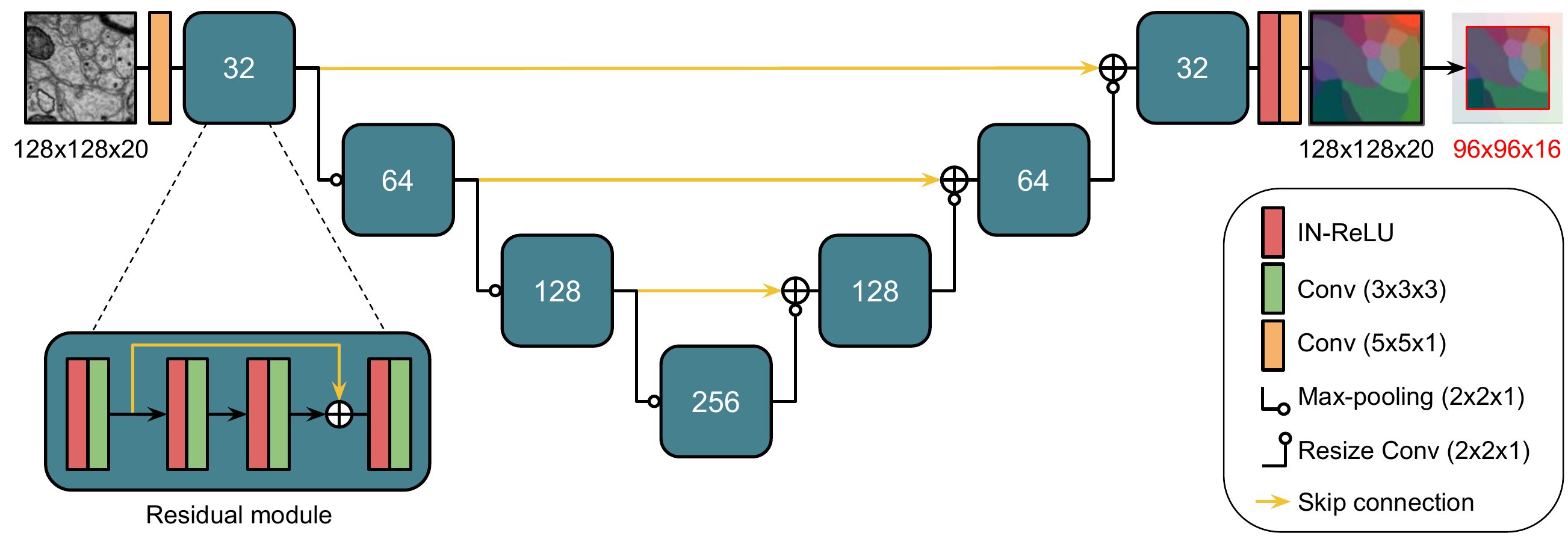}
\caption{Embedding net architecture. The number inside each of the residual modules represents the width (number of feature maps) of the module. Here the dimensionality of output embeddings is omitted for brevity and only spatial dimensions are displayed. Abbreviations: Instance Normalization (IN), rectified linear unit (ReLU), convolution (Conv).}
\label{fig:embedding_net}
\end{center}
\end{figure*}

\begin{figure*}[!ht]
\begin{center}
\includegraphics[width=0.95\textwidth]{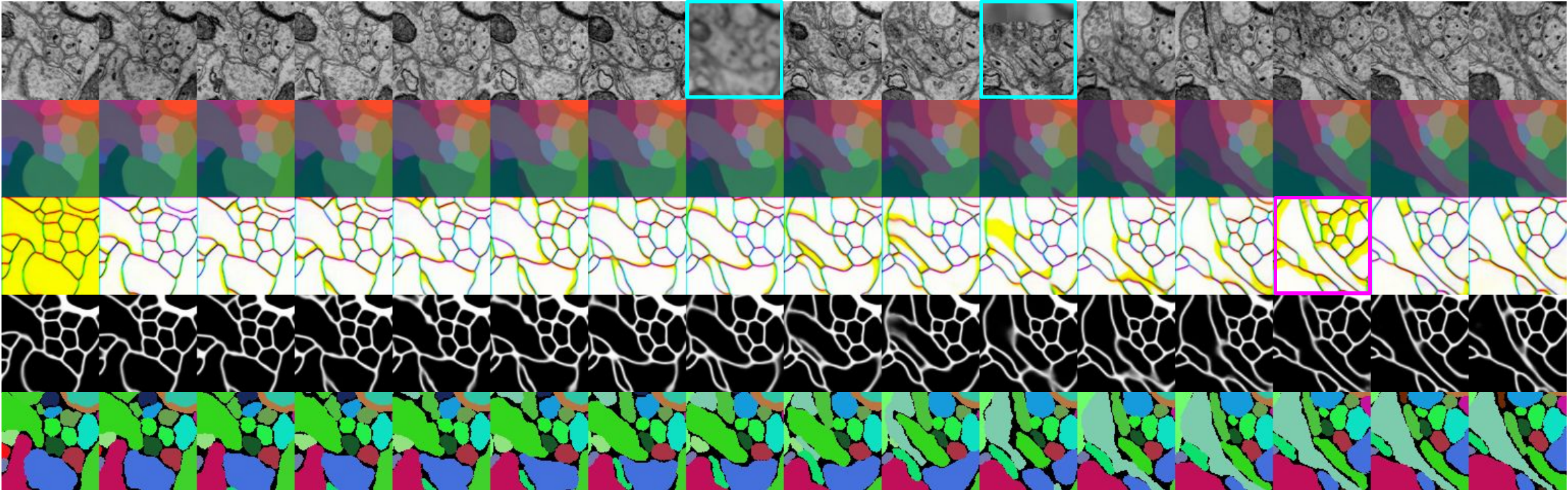}
\caption{Visualization of an example training patch. Each row is a flattened version of $96\times96\times16$ image patch. From top to bottom, each row shows (1) EM images, (2) dense voxel embeddings, (3) nearest neighbor metric graph derived from the embeddings, (4) predicted background mask, and (5) ground truth segmentation. Yellow-colored regions in the visualization of nearest neighbor metric graph (third row) indicate disconnectivity in $z$-direction. In this particular training example, one full and one partial simulated out-of-focus sections (cyan boxes) and one simulated misalignment (magenta box) were injected.} \label{fig:example_training_patch}
\end{center}
\end{figure*}

\begin{figure*}[!ht]
\begin{center}
\includegraphics[width=0.95\textwidth]{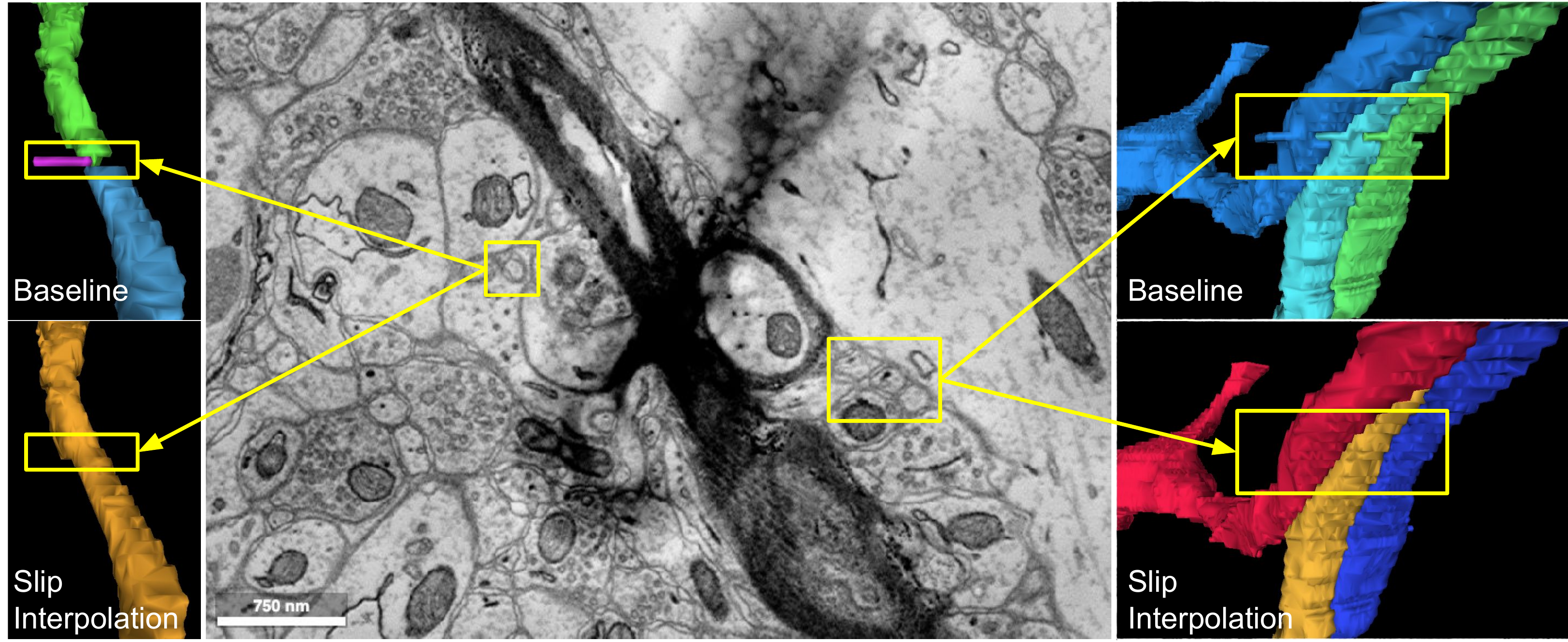}
\caption{Effect of slip interpolation. A mild fold (dark shade over the myelinated axon) in the test set causes a slip-type misalignment. {\bf Left:} a thin axon is broken at the slip misalignment (top), whereas the baseline net trained with slip interpolation produces a smoothly interpolated output and heals the split error (bottom). {\bf Right:} three abutting axons are not broken but affected by the slip misalignment (top), whereas the baseline net trained with slip interpolation produces smoothly interpolated segments (bottom).} \label{fig:slip_interpolation}
\end{center}
\end{figure*}

\begin{figure*}[!ht]
\begin{center}
\includegraphics[width=0.95\textwidth]{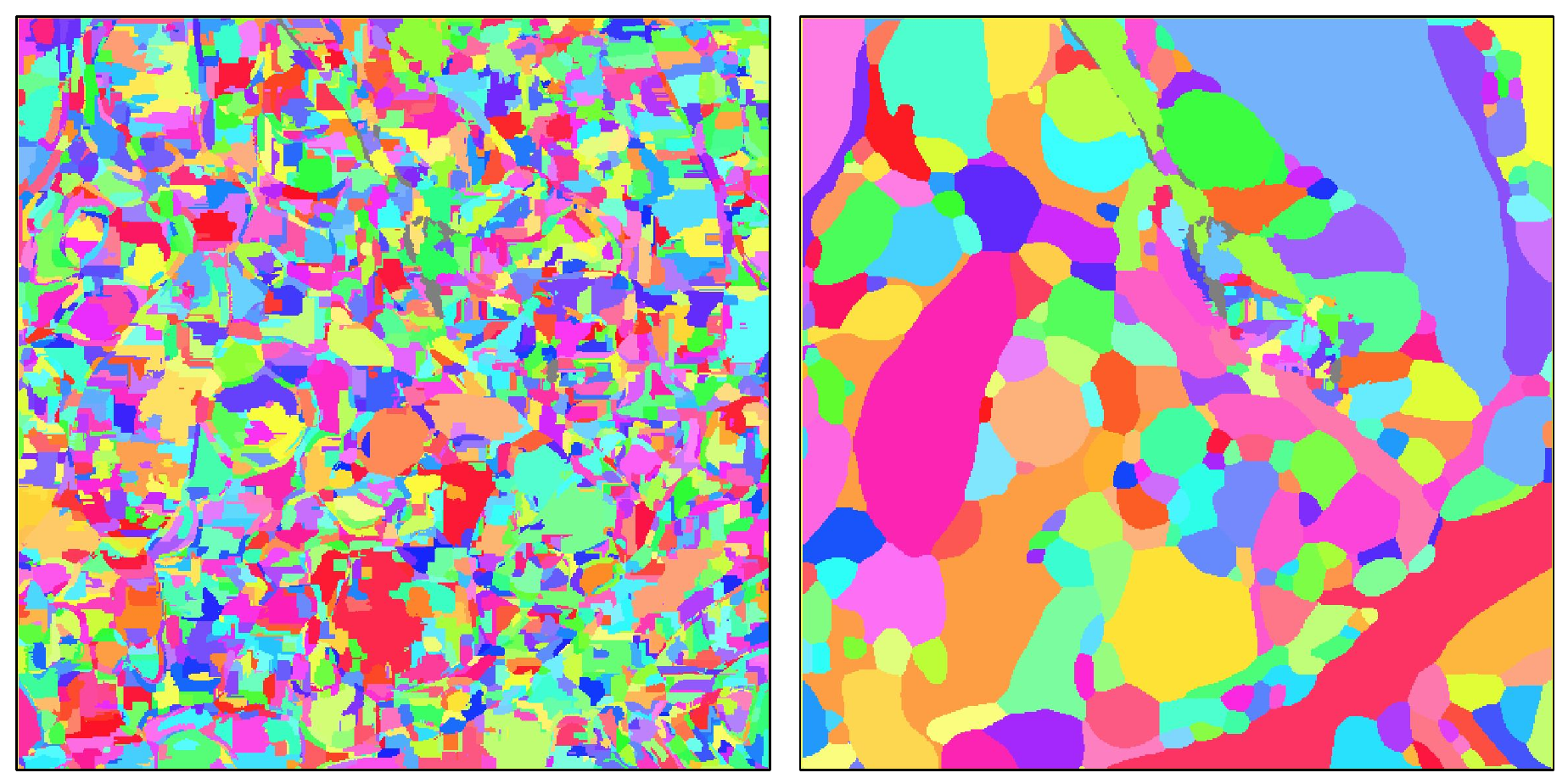}
\caption{Watershed oversegmentation for the baseline. (a) To obtain initial oversegmentation, we used $T_\text{min} = 1\%$, $T_\text{max} = 99\%$, $T_\text{size} = (\theta_\text{size}, 1\%)$, $T_\text{dust} = \theta_\text{size}$, where $\theta_\text{size} = 600$ at the original resolution ($6\times6\times29$ nm$^3$) and $\theta_\text{size} = 150$ at the $2\times$ in-plane-downsampled resolution ($12\times12\times29$ nm$^3$). (b) Final segmentation can be obtained by greedily agglomerating supervoxel pairs whose \emph{agglomeration score} (max or mean affinity) are higher than a tunable threshold.}
\label{fig:oversegmentation}
\end{center}
\end{figure*}

\begin{figure*}[!ht]
\begin{center}
\includegraphics[width=0.95\textwidth]{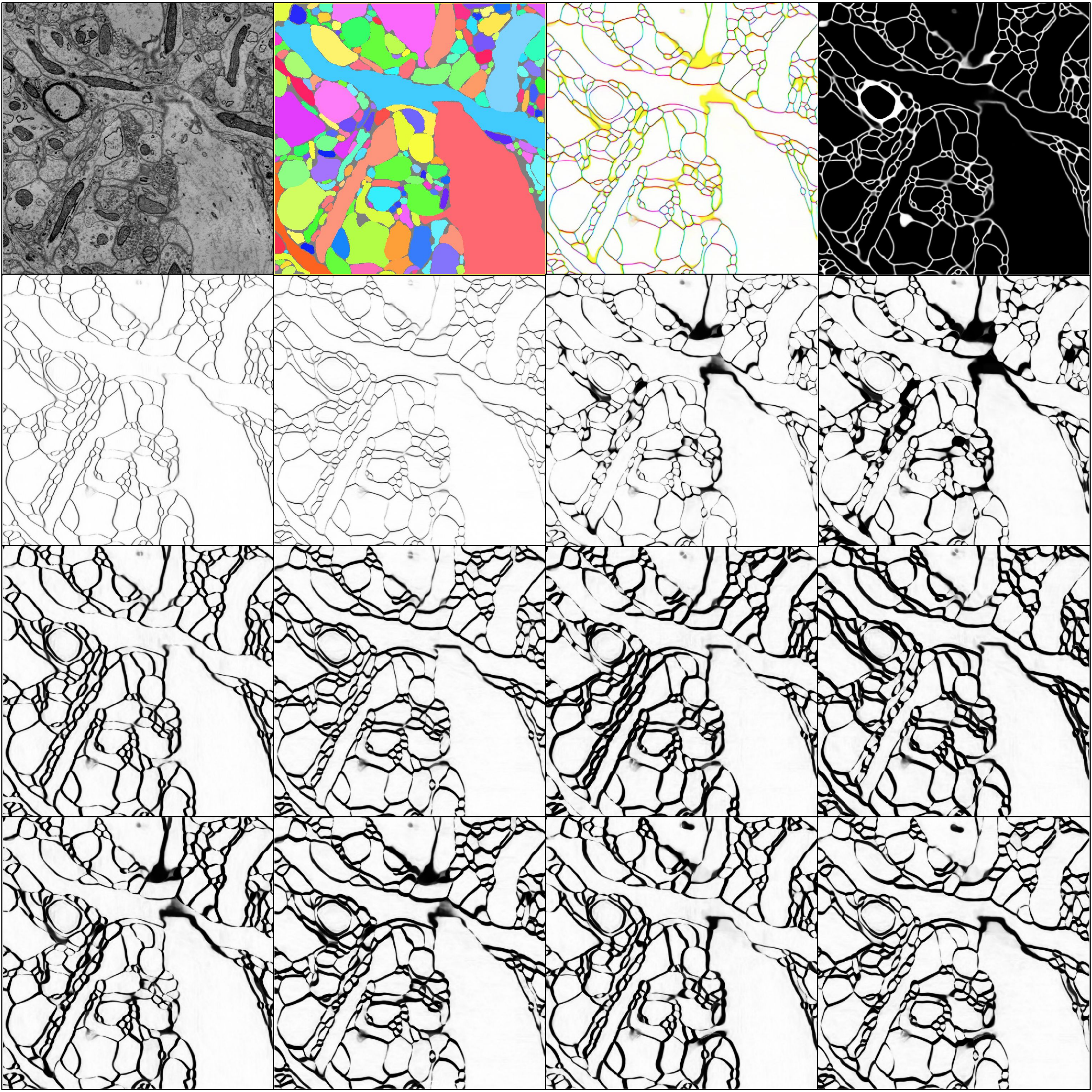}
\caption{Metric graph on short and long-range edges as input to the Mutex Watershed. Top row shows from left to right (1) input image from the validation set, (2) ground truth segmentation, (3) RGB visualization of the nearest neighbor metric graph, and (4) predicted background mask. For the Mutex Watershed, we used three nearest neighbor \emph{attractive} edges and nine long-range \emph{repulsive} edges. Each edge yields an affinity map with $(x,y,z)$ offset of (-1,0,0), (0,-1,0), (0,0,-1), (0,0,-2) on the second row, (-5,0,0), (0,-5,0), (-5,-5,0), (-5,5,0) on the third row, and (-5,0,-1), (0,-5,-1), (-5,0,1), (0,-5,1) on the bottom row. We used sufficient image padding when computing the affinities near the dataset edge. Each affinity map is obtained by stitching and blending the patch-wise affinity maps derived from the patch-wise dense voxel embeddings.}
\label{fig:edge_neighborhood_structure}
\end{center}
\end{figure*}

\begin{figure*}[!ht]
\begin{center}
\includegraphics[width=1.0\textwidth]{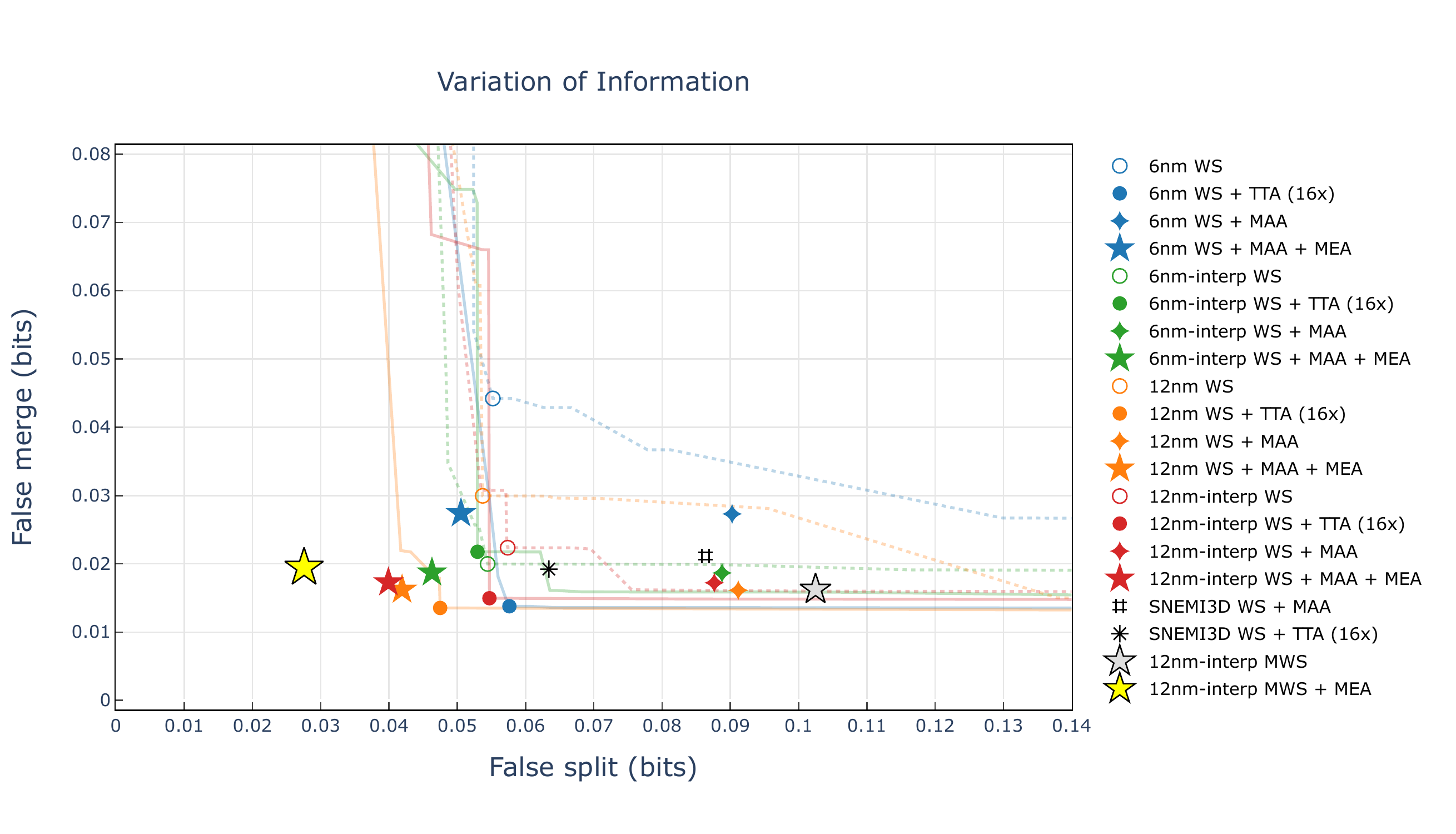}
\caption{Full merge-split plot on the test set. For comparison, we have included the SNEMI3D submission results of Lee \textit{et al.} (2017). Asterisk (*) represents the top entry of the SNEMI3D leaderboard by Lee \textit{et al.} (2017) based on $16\times$ test-time augmentation (adapted Rand error of 0.0249 reported by the SNEMI3D leaderboard). Pound sign (\#) represents the SNEMI3D submission result by Lee \textit{et al.} (2017) based on mean affinity agglomeration (adapted Rand error of 0.0314 reported by the SNEMI3D leaderboard). Abbreviations: watershed (WS), mean affinity agglomeration (MAA), test-time augmentation (TTA), mean embedding agglomeration (MEA), slip interpolation (interp), original image resolution (6 nm), 2$\times$ in-plane-downsampled image resolution (12 nm).} \label{fig:AC3_test_merge_split}
\end{center}
\end{figure*}

\begin{figure*}[!ht]
\begin{center}
\includegraphics[width=0.5\textwidth]{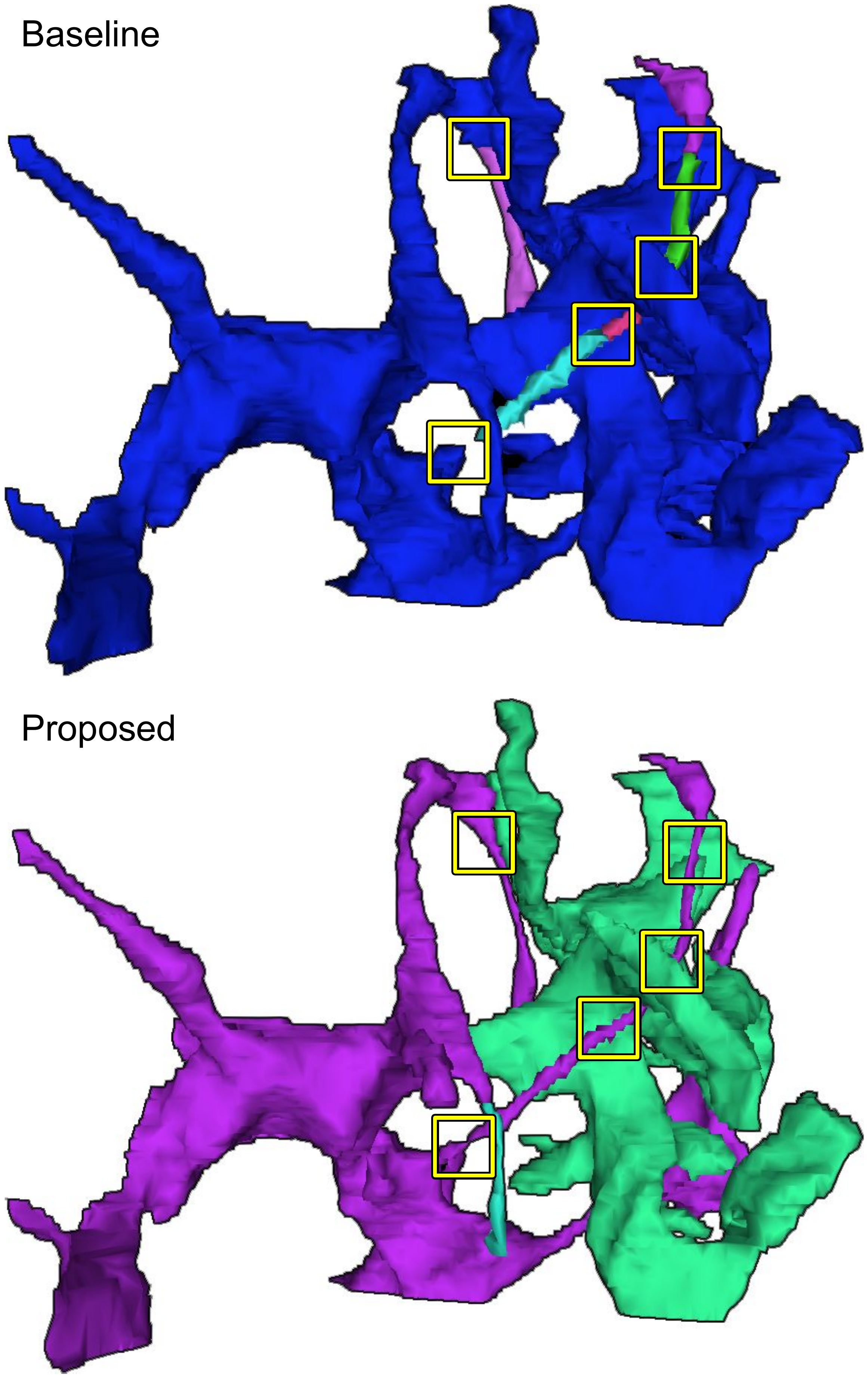}
\caption{Dense voxel embeddings bring substantial accuracy gain for very thin glia processes. Shown here is a complex glia (putative astrocyte) from the extra test volume E1. The baseline segmentation (top) has systematic split errors concentrated on the very thin glia processes (yellow boxes), whereas our proposed method successfully extends them. However, the proposed method made a couple of split errors (green and cyan segments) due to the conservative repulsive constraints put by the Mutex Watershed on self-contact. Here mean embedding agglomeration failed to resolve a couple of self-contact split errors, mainly due to the heuristic's failure in detecting them.} \label{fig:thin_object_comparison_glia}
\end{center}
\end{figure*}

\begin{figure*}[!ht]
\begin{center}
\includegraphics[width=0.8\textwidth]{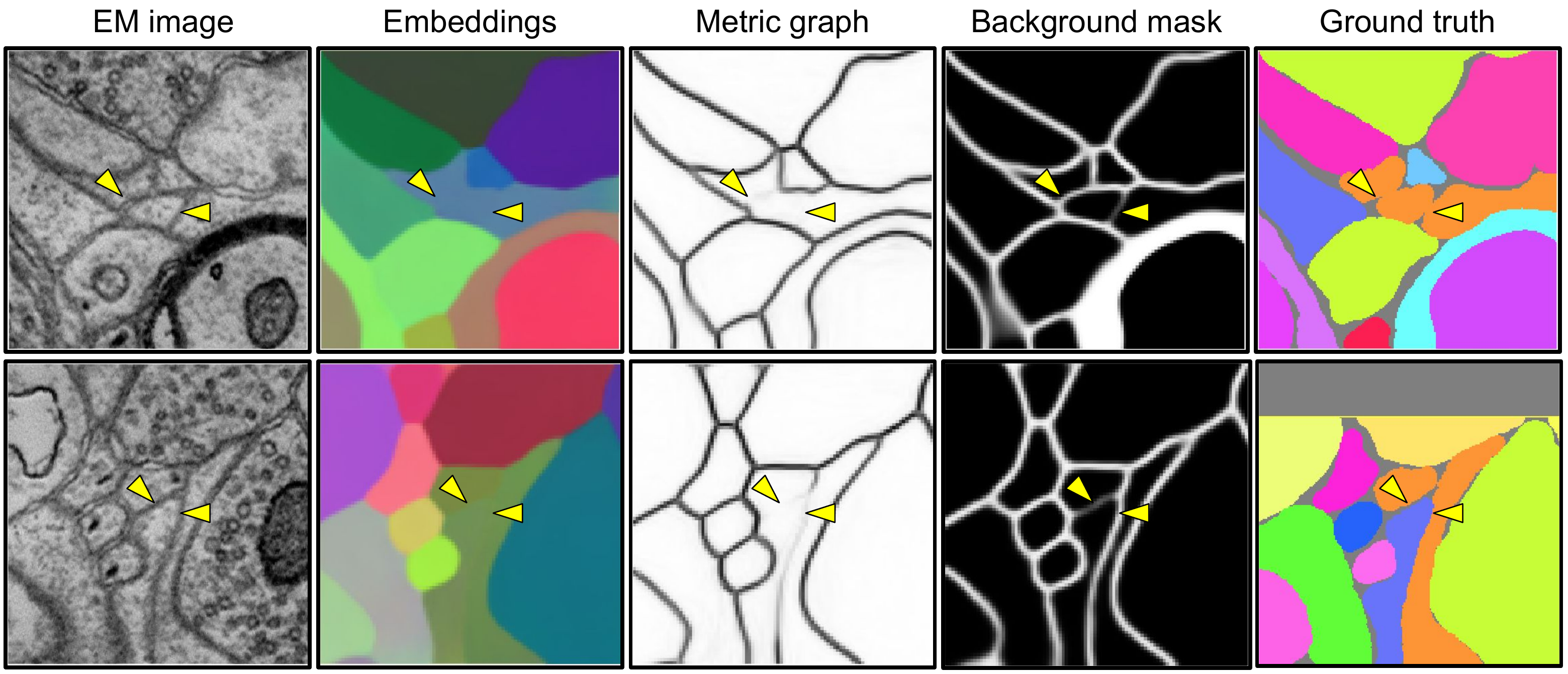}
\caption{The embedding net is confused by complex glia self-contacts in the training set (AC3). {\bf Top:} the embedding net ``memorizes'' the glia self-contacts in the training set (orange object, last column) by assigning uniform embeddings across the self-contacts (yellow arrowheads, second column), effectively erasing boundaries in the nearest neighbor metric graph (third column). {\bf Bottom:} the embedding net makes a mistake on a similar-looking location in the training set where the contacts are between two distinct glia this time (yellow arrowheads, last column). As can be seen here, glia with complex morphology (putative astrocytes) make numerous self-contacts that are not properly separated by background voxels in the ground truth annotation. This becomes a significant source of noise during training, systematically compromising the embedding net's performance around glia, even in the training set. To visualize metric graph (third column), we used $\min(a_x, a_y)$ where $a_x$ and $a_y$ are nearest neighbor metric-derived affinities in $x$ and $y$ directions, respectively.} \label{fig:glial_confusion}
\end{center}
\end{figure*}

\begin{figure*}[!ht]
\begin{center}
\includegraphics[width=0.9\textwidth]{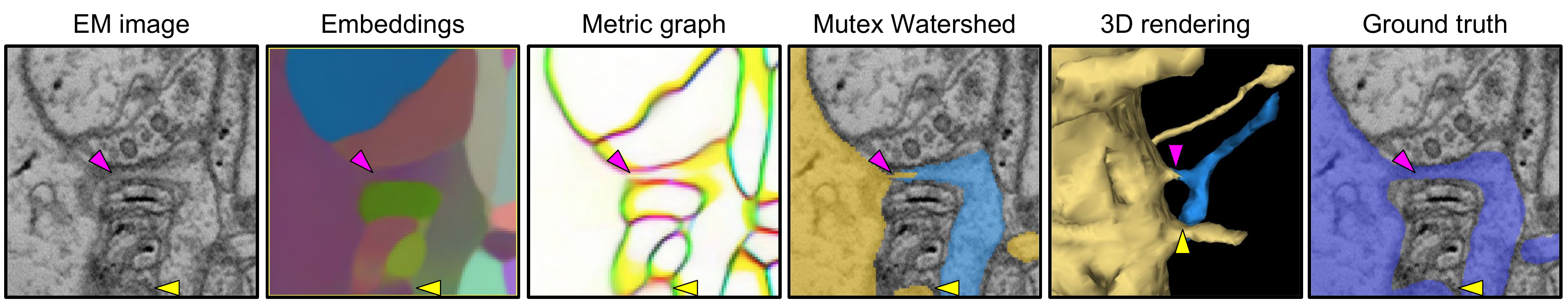}
\caption{Self-contact within local patch. A dendritic spine that bends back and makes a self-contact (yellow arrowheads) within the field of view of the embedding net, taken from the training set (AC3). The embedding net fails to assign uniform vectors across the dendritic shaft and spine (magenta arrowheads), despite that the net was trained on this example.} \label{fig:self_touching_within_patch}
\end{center}
\end{figure*}

\begin{figure*}[!ht]
\begin{center}
\includegraphics[width=0.9\textwidth]{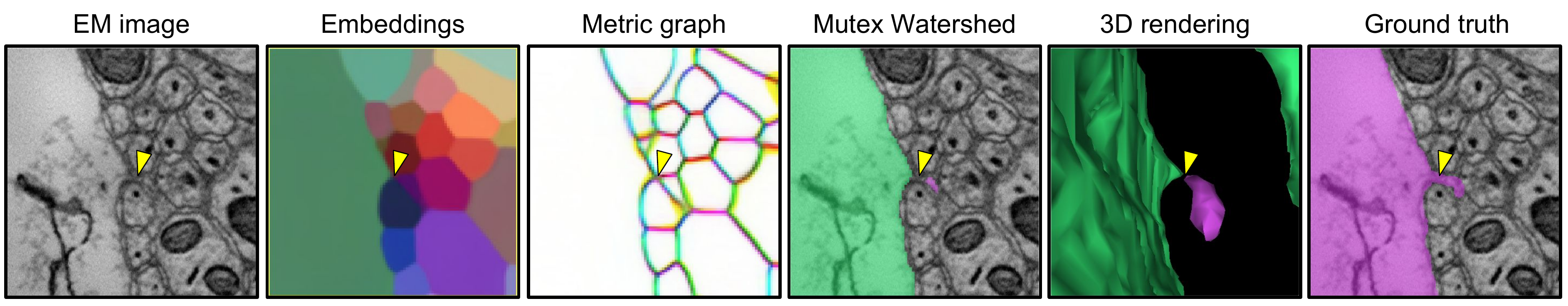}
\caption{A tiny broken spine in the validation set (AC3). The embedding net assigns completely distinct vectors to the dendritic shaft and the tiny spine (yellow arrowheads).} \label{fig:tiny_spine}
\end{center}
\end{figure*}

% \end{document}

\end{document}